\documentclass[letterpaper, 10 pt, conference]{ieeeconf}  %

\IEEEoverridecommandlockouts                              %

\usepackage{hyperref} 
\usepackage{mathrsfs}
\usepackage{amsfonts}
\usepackage{amsmath}
\usepackage[flushleft]{threeparttable}
\usepackage{tablefootnote}
\usepackage{multirow}
\usepackage{graphicx}
\usepackage{subfigure}
\let\labelindent\relax
\usepackage{enumitem}
\usepackage{hanging}
\usepackage{color}
\usepackage{tikz}
\usetikzlibrary{calc,arrows,decorations.markings}
\usepackage{balance}

\def\bmatrix#1{\left[ \begin{matrix} #1 \end{matrix} \right]}  %
\newcommand{\bfR}{\ensuremath{\textbf{R}}}
\newcommand{\bfT}{\ensuremath{\textbf{T}}}
\newcommand{\bfk}{\ensuremath{\textbf{k}}}
\newcommand{\bft}{\ensuremath{\textbf{t}}}
\newcommand{\R}{\mathbb{R}}  %
\newcommand{\calT}{\ensuremath{{\cal T}}}

\usepackage{booktabs}  %

\usepackage{cuted}
\usepackage{capt-of}

\title{\LARGE \bf
Multi-View Fusion for Multi-Level Robotic Scene Understanding}

\author{Yunzhi Lin$^{1,2}$, Jonathan Tremblay$^{1}$, Stephen Tyree$^{1}$, Patricio A.  Vela$^{2}$ and Stan Birchfield$^{1}$, 
\\ $^{1}$NVIDIA: {\tt\small \{jtremblay, styree, sbirchfield\}@nvidia.com}
\\ $^{2}$Georgia Institute of Technology: {\tt\small \{yunzhi.lin, pvela\}@gatech.edu}
\thanks{Work was completed while the first author was an intern at NVIDIA.}%
}

\begin{document}

\twocolumn[{%
\renewcommand\twocolumn[1][]{#1}%
\maketitle

\begin{center}
    \centering
    \begin{tabular}{ccc}
    {\hspace{-0.09in}\includegraphics[width=0.33\textwidth,clip=true,trim=0in 0in 0in 0in]{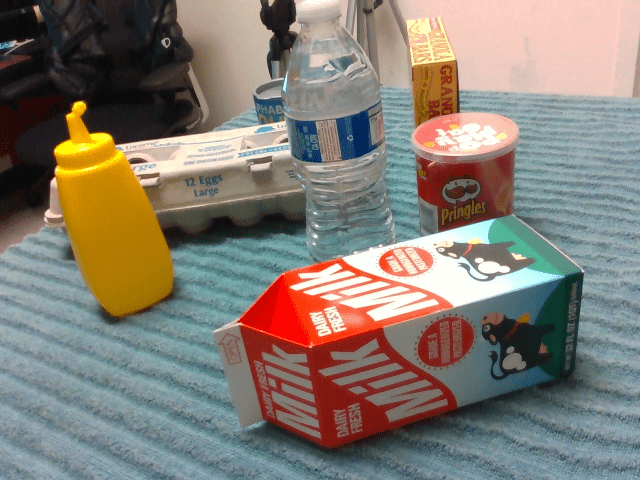}} 
    &
    {\hspace{-0.15in}\includegraphics[width=0.33\textwidth,clip=true,trim=
    1.8in 2.25in 2.0in 0.9in]{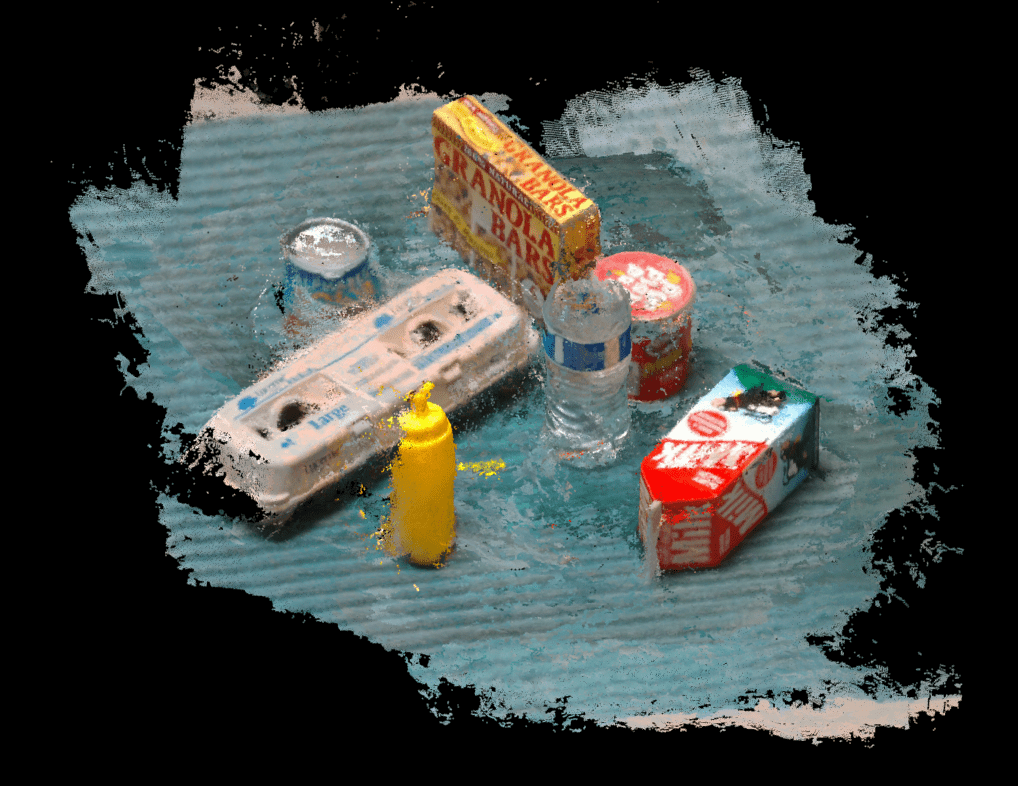}} 
    &    
    {\hspace{-0.15in}\includegraphics[width=0.33\textwidth,clip=true,trim=0.3in 2.35in 3.0in 0.4in]{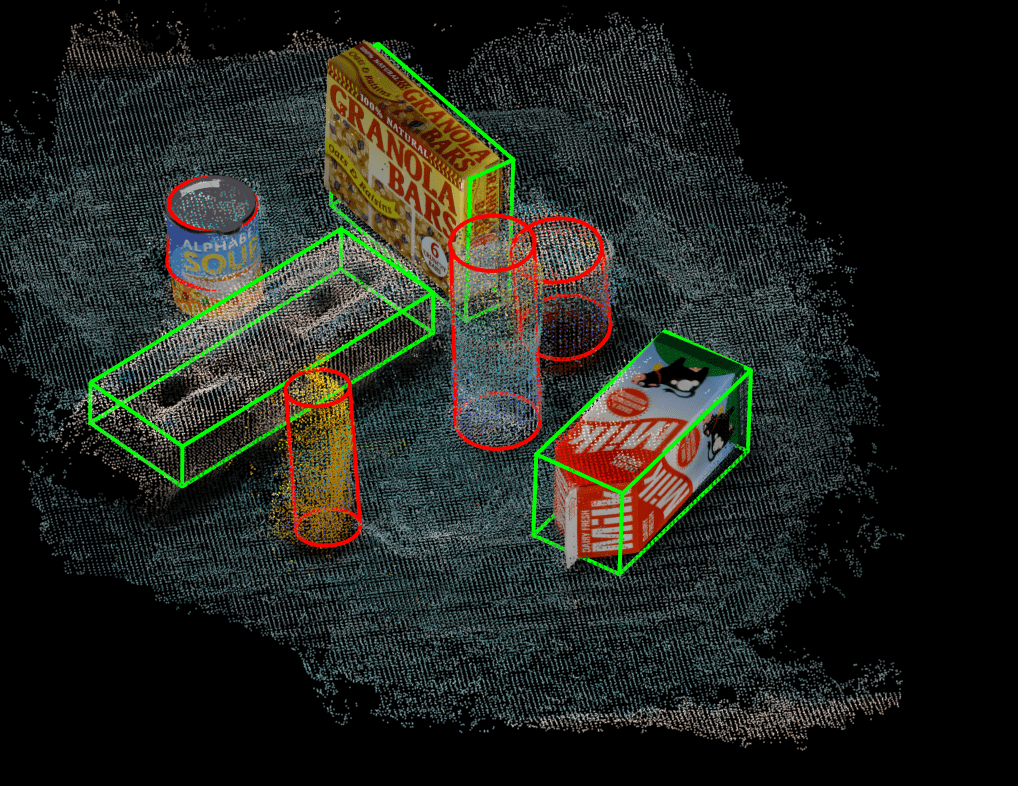}}
    \end{tabular}
    \captionof{figure}{{\sc Left:} Input to the system is a sequence of RGB images from a camera-in-hand.  {\sc Middle:}  Reconstructed point cloud.  {\sc Right:} Multi-level scene representation, including 3D point cloud (downsampled for better visualization), primitive shapes (cylinders and cuboids), and known object CAD models (granola bars, milk, and soup can).\label{fig:abstract_demo}}
\end{center}%
}]

\begin{abstract}
We present a system for multi-level scene awareness for robotic manipulation.
Given a sequence of camera-in-hand RGB images, the system calculates three types of information:  1) a point cloud representation of all the surfaces in the scene, for the purpose of obstacle avoidance; 2) the rough pose of unknown objects from categories corresponding to primitive shapes (e.g., cuboids and cylinders); and 3) full 6-DoF pose of known objects.
By developing and fusing recent techniques in these domains, we provide a rich scene representation for robot awareness.
We demonstrate the importance of each of these modules, their complementary nature, and the potential benefits of the system in the context of robotic manipulation.\makeatletter{\renewcommand*{\@makefnmark}{}\footnotetext{Work was completed while the first author was an intern at NVIDIA.  Video is at \url{https://youtu.be/FuqMxuODGlw}.}\makeatother}
\end{abstract}

\section{Introduction}

Scene awareness, or scene understanding, is critical for a robotic manipulator to interact with an environment.  
A robot must know both \emph{where} surfaces are located in the scene, for obstacle avoidance, as well as \emph{what} objects are in the scene for grasping and manipulation.
Some objects may be known to the robot and relevant to the task at hand, while others may only be recognizable by their general category or affordance properties.
Despite the tremendous progress made in the computer vision community on solving problems such as 3D reconstruction~\cite{schonberger2016structure,mccormac2018dv:fusionpp,gu2020cascade,mildenhall2020eccv:nerf} and object pose estimation~\cite{hinterstoisser2012accv:linemod,peng2019cvpr:pvnet,zakharov2019iccv:dpod,hu2019cvpr:segdriv6d,hodan2018eccv:bop,
rennie2016ral:rutgersapc,kaskman2019arx:homebrew,park2019iccv:pix2pose,tremblay2018corl:dope,xiang2018rss:posecnn}, existing deployed robotic manipulators have  limited, if any, perception of their surroundings. 

To overcome this limitation, we argue that a robotic manipulator needs three levels of understanding:
\begin{itemize}[leftmargin=*]
    \item \emph{Generic surfaces.}  As the robot moves within the workcell, it is important to avoid unintended collisions to maintain safe operation.  Therefore, it must be aware of obstacles nearby, whether or not they are manipulable.
    \item \emph{Known categories / affordances.}  Some of these surfaces will be objects that are manipulable.  
    For many such objects it may be sufficient to simply recognize the category to which the object belongs, or some affordance properties.
    In this work, we mainly find objects whose shape is roughly cylindrical or cuboidal.
    \item \emph{Known objects.}  Some of these objects may be known beforehand.  For example, oftentimes a robot is deployed in a workcell to interact with a small set of known objects for a specific task.  For such objects, it should be possible to infer their full 6-DoF poses for rich manipulation.
    
\end{itemize}

We present a system that integrates these three levels of understanding, see Fig.~\ref{fig:abstract_demo}.
Unlike existing approaches to integrating object-level perception and robotic manipulation~\cite{kappler2017arx:motiongen,wada2020cvpr:morefusion,sucar2020arxiv:nodeslam,murali2020icra:tdomic,zhenjia2020corl:dsr}, which rely on depth sensing, our system relies on RGB images as input. In the case of a static scene, 3D information can be recovered with multi-view RGB images via triangulation from correspondences. Color cameras generally operate at high resolution and therefore yield potentially detailed scene information. 
Moreover, RGB is often needed to correct errors in depth measurements, like those due to transparent surfaces~\cite{sajjan2020icra:cleargrasp}, and it also has a larger working range.
In recent years, RGB processing has experienced significant growth in capability, including depth estimation~\cite{fu2018deep}, flow field prediction~\cite{mildenhall2020eccv:nerf}, and object pose estimation~\cite{tremblay2018corl:dope}. 

\begin{figure*}[t!]
  \vspace*{0.075in}
  \centering
  \begin{tikzpicture}[inner sep = 0pt, outer sep = 0pt]
    \node[anchor=south west] at (0in,0in)
      {{\includegraphics[width=\textwidth,clip=true,trim=0in 0in 0in 0in]{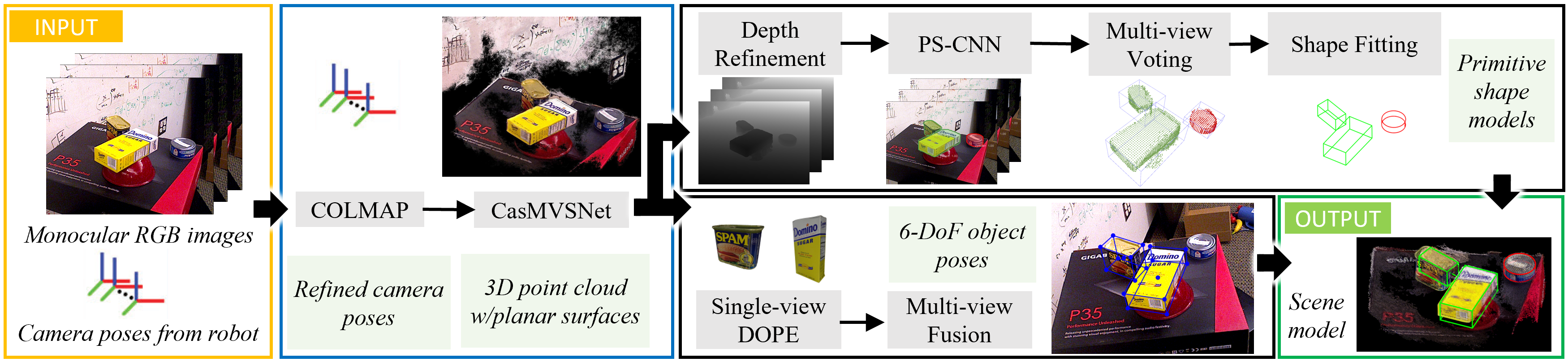}}};
      \node[anchor=south] at (2.10in,1.63in) 
	    {\footnotesize \S\ref*{approach:casmvsnet}: \sc Dense 3D reconstruction};
      \node[anchor=south west] at (3.1in,1.63in) 
		{\footnotesize \S\ref*{approach:primShape}: \sc Multi-view primitive shape segmentation and fitting};
	 \node[anchor=north west] at (3.1in,0in) 
        {\footnotesize \S\ref*{approach:pose fusion}: \sc Multi-view object pose estimation};
  \end{tikzpicture}
  \caption{Processing pipeline.  Given a sequence of RGB images from a
  camera-in-hand, and the corresponding approximate camera poses from forward
  kinematics, our system provides multi-level scene awareness for robotic
  manipulation. The three components of the system are dense 3D
  reconstruction, primitive shape fitting, and known object pose
  estimation. The final output is an integrated scene model (zoom for the best view), including a 3D point cloud of the scene, primitive shape models, and 6-DoF object poses.
  \label{fig:pipeline}}%
  \vspace*{-3.0ex}
\end{figure*}
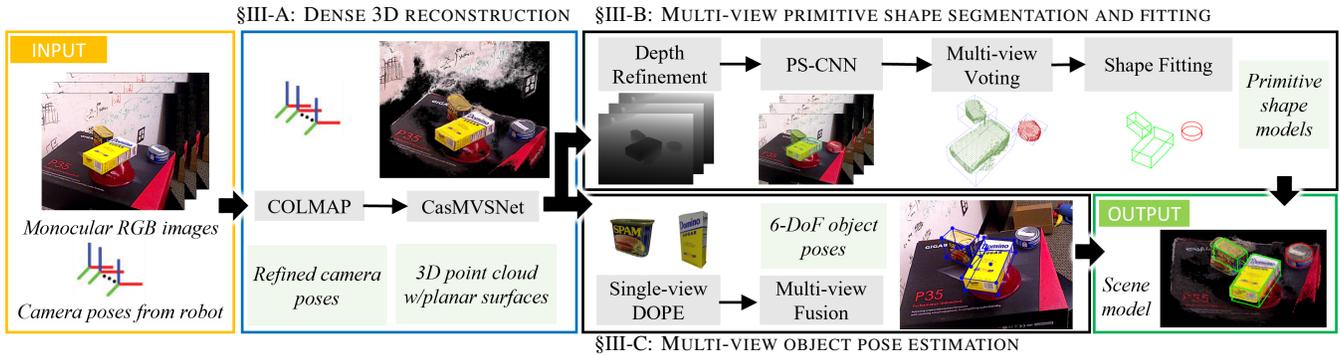

Our system scans a scene using an RGB eye-in-hand camera, and processes the image sequence to generate a multi-level representation of the scene.
Specifically, the system consists of three components:  1) dense 3D reconstruction using 
COLMAP~\cite{schonberger2016structure} and CasMVSNet~\cite{gu2020cascade}, with a novel post-processing step to yield high-quality depth maps; 2) a primitive shape network that works on a large cluttered environment based on multi-view segmentation results to generate solid parametric model fits of scene objects; and 3) a multi-view object pose estimator based on the concept of late fusion.

Our work makes the following contributions: 
\begin{itemize}[leftmargin=*]
    \item Multi-level scene understanding for robotic manipulation, including dense 3D  reconstruction for obstacle avoidance, shape estimation and fitting of objects with primitive shapes, and full 6-DoF pose estimation of known object instances.
    \item Multi-view primitive shape fitting algorithm consuming virtual depth maps from RGB-based reconstruction and capable of handling a large cluttered environment.
    \item Multi-view RGB-based object pose estimator with a voting mechanism and Levenberg-Marquardt based refinement.
    \item Evaluation of these modules on real RGB camera-in-hand sequences, using a new dataset we captured and annotated called HOPE-video.  Results demonstrate the improved performance that arises from our multi-view extensions.
\end{itemize}

\section{Related Work}

One of the more successful systems employing object-level perception for robotic manipulation is the MOPED system~\cite{romea2011ijrr:moped}, which uses RGB-based perception to identify the locations of a large number of objects for grasping.
More recent works rely on RGBD or depth perception for scene interpretation, obstacle recognition, and object manipulation or grasping~\cite{kappler2017arx:motiongen,murali2020icra:tdomic}.
Both MoreFusion~\cite{wada2020cvpr:morefusion} and NodeSLAM~\cite{sucar2020arxiv:nodeslam} present promising RGBD-based approaches complementary to ours.

{\bf RGB-based 3D dense reconstruction.} 
Typical RGB-based 3D dense reconstruction systems consist of two parts: structure-from-motion (SfM) \cite{schonberger2016structure} and multi-view stereo (MVS)~\cite{schonberger2016pixelwise}. Open-source SfM methods \cite{schonberger2016structure,moulon2016openmvg,theia-manual} work well in textured scenes, given coarse initial camera pose estimations from the robotic arm. Meanwhile, traditional multi-view stereo algorithms \cite{lhuillier2005quasi,furukawa2009accurate,tola2012efficient} have difficulty recovering the details of small objects or require large processing times. 
Learning-based multi-view stereo approaches demonstrate superior performance and wide generality on different scenarios. MVSNet~\cite{yao2018mvsnet} builds the 3D cost volume on the reference camera frustum via the differentiable homography warping, and 3D CNNs are applied for cost regularization and depth regression. In followup work, CasMVSNet~\cite{gu2020cascade} improves the accuracy and computational speed by integrating a cascade cost volume of high-resolution.

{\bf Primitive shapes for robotic manipulation.} 
The idea of using primitive shapes for robotic grasping of unknown objects within known categories relies on generating grasp configurations by modeling an
object as a set of shape primitives~\cite{miller2003automatic}. Past research explores different shape primitive approaches, including a graph representation \cite{aleotti20123d}, superquadric fitting \cite{Goldfeder_Allen_Lackner_Pelossof_2007,vezzani2017grasping,xia2018reasonable}, and box surface approximation \cite{huebner2008selection}. Recent research \cite{lin2020using} has achieved state-of-the-art performance based on a single-view segmentation pipeline. 
Our approach expands this work to handle more cluttered scenes by a novel data generation procedure and multi-view voting procedure.

{\bf Multi-view object pose fusion.} 
Recent object-level SLAM systems \cite{salas2013slam++,mccormac2018fusion++,sucar2020arxiv:nodeslam} that jointly optimize the poses of detected objects and cameras have presented promising results. 
More traditional, decoupled systems, process each view individually then select consistent hypotheses for global refinement given camera poses. 
Collet et al.~\cite{collet2010efficient} calculate the minimal sum of reprojection errors of correspondences across all images after mean-shift clustering. 
Erkent et~al.~\cite{erkent2016integration} formulate a probabilistic framework to fuse pose estimates from different views. 
Sock et al.~\cite{sock2017multi} choose the representative hypothesis based on subtractive clustering and confidence score. 
Li et al.~\cite{li2018unified} perform hypothesis voting based on approximated average distances between hypotheses. 
In contrast, our proposed method leverages the single-view pose estimation method DOPE \cite{tremblay2018corl:dope} by incorporating bounding box keypoint predictions with weighted sum of reprojection error.

{\bf Multi-level scene understanding for manipulation.} 
Multi-level scene understanding aims to provide a unified scene representation for advanced manipulation tasks such as collision-free grasping or object rearrangement. 
Previous approaches build a hierarchical representation based on edge and texture information \cite{popovic2011grasping} or present a task-oriented grasp algorithm based on affordance-region segmentation and geometric grasp model searching \cite{detry2017task}. Such approaches generate grasp configurations without object-level understanding. 
The most relevant work to the proposed system is that of Bohg et al.~\cite{bohg2011multi}, which divides the scene into different layers, including an occupancy grid, recognition and pose estimation for known objects, and object shape estimation based on shape surface plane model for unknown objects. 
Inspired by this work, our approach aims to operate in a more cluttered environment by making full use of the recent advances in deep learning-based computer vision.

\section{Approach}

Our system leverages three modules to produce  three different levels of representation for robotic manipulation.
It assumes a camera mounted on a robot arm captures multiple views of a
scene and registers the camera pose at each capture. 
Fig.~\ref{fig:pipeline} describes the general workflow: 3D reconstruction, primitive shape fitting, and 6-DoF pose estimation of known objects.

\subsection{Multi-view stereo for 3D dense reconstruction}
\label{approach:casmvsnet}

Dense 3D scene reconstruction is needed for obstacle avoidance and as input to the other modules.
We use a two-step process that invokes COLMAP~\cite{schonberger2016structure} to refine the camera poses obtained from the robot, as shown in Fig.~\ref{fig:pipeline}. This helps to decrease the camera pose errors caused by robot forward kinematic discrepancies, synchronization issues, {\em etc}. The motion prior from the robot has sub-degree accuracy and mainly serves to speed up the processing time for COLMAP. Given COLMAP refined camera poses, the second step relies on CasMVSNet~\cite{gu2020cascade}, a deep-learning-based multi-view stereo method, to provide a dense, colored 3D point cloud. This multi-view stereo method leverages a feature pyramid geometric encoding that uses coarse-to-fine processing.

\subsection{Multi-view primitive shape segmentation and fitting \label{approach:primShape}}

Given the point cloud output from the previous process, we seek to find all the possible graspable objects through a 
shape completion algorithm. 
For this we use our recent PS-CNN method~\cite{lin2020using}, which decomposes common household objects into one or more primitive shapes for grasping, using a single depth image. However, its success relies on a system identification prerequisite that the simulation and the real world share the same environment setup, e.g., the camera's setup.
It also does not focus on a large cluttered environment and is sensitive to noisy backgrounds. To adapt and extend the method for the proposed system, we introduce the following improvements.

\begin{figure}[t]
  \vspace*{0.075in}
  \centering
  \begin{tikzpicture} [outer sep=0pt, inner sep=0pt]
  \scope[nodes={inner sep=0,outer sep=0}] 
  \node[anchor=north west] (a) at (0in,0in) 
    {\includegraphics[height=3.5cm,clip=true,trim=1in 0in 0in 0in]{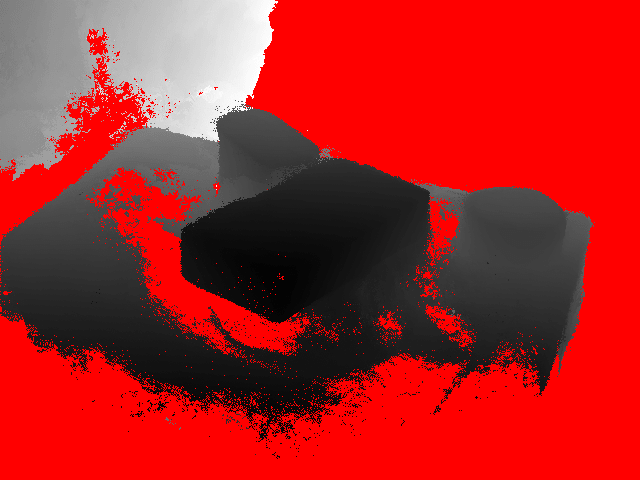}}; 
  \node[anchor=north west, xshift=0.1cm] (b) at (a.north east)
    {{\includegraphics[height=3.5cm,clip=true,trim=1in 0in 0in 0in]{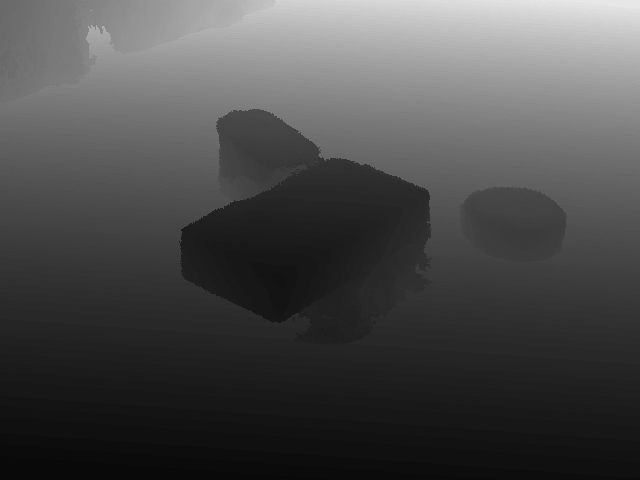}}}; 
  \endscope

  \end{tikzpicture}
  \vspace*{-1.0ex}
  \caption{{\sc Left:} Raw, noisy virtual depth map from RGB-based reconstruction. Red indicates missing pixels. {\sc Right:} Refined virtual depth map.\label{fig:approach_depth_refinement} }
  \vspace*{-3.0ex}
\end{figure}

\subsubsection{Depth refinement}

PS-CNN~\cite{lin2020using} expects a high-quality depth images from a depth sensor, whereas our system must utilize virtual depth images rendered from the reconstructed point cloud.
To remove undesirable artifacts, we first denoise the resulting point cloud, then apply RANSAC to identify tabletop plane parameters, after which double thresholding removes and replaces the tabletop points without affecting the objects on the table.
The resulting point cloud is projected onto the image plane to yield a virtual depth map, with region connectivity-based denoising, temporal averaging, and spatial median filtering. Finally, the virtual tabletop plane is re-introduced to fill the missing pixels.
Fig.~\ref{fig:approach_depth_refinement} shows the raw and refined virtual depth images, illustrating the improvement in quality.

\subsubsection{Data generation}

\begin{figure}[t]
  \vspace*{0.075in}
  \centering
  \begin{tikzpicture} [outer sep=0pt, inner sep=0pt]
  \scope[nodes={inner sep=0,outer sep=0}] 
  \node[anchor=north west] (a) at (0in,0in) 
    {\includegraphics[height=3.5cm,clip=true,trim=1in 0in 0in 0in]{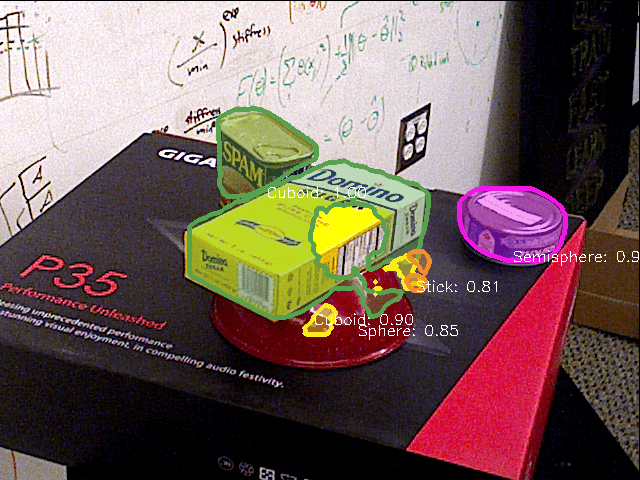}}; 
  \node[anchor=north west, xshift=0.1cm] (b) at (a.north east)
    {{\includegraphics[height=3.5cm,clip=true,trim=1in 0in 0in 0in]{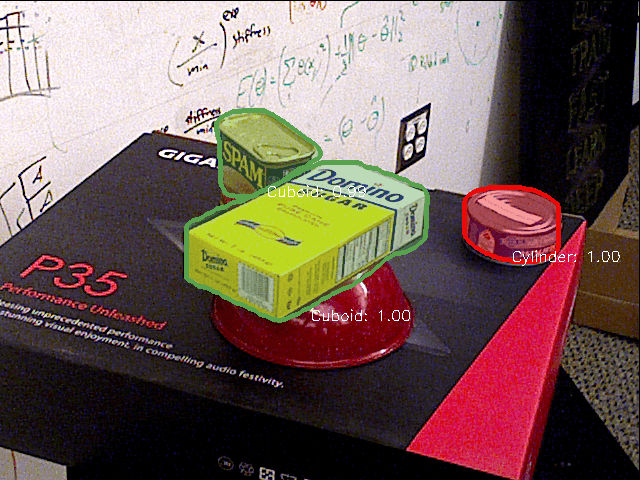}}}; 
  \endscope

  \end{tikzpicture}
  \vspace*{-1.0ex}
  \caption{Primitive-shape segmentation using the system of~\cite{lin2020using} (left) and ours (right) on the same noisy virtual depth image (RGB, not used, is shown for viewing clarity).  Our network yields cleaner results on the sugar box, and the correct label for the small can.} 
  \label{fig:approach_seg}
\end{figure}

Since PS-CNN \cite{lin2020using} is trained on scenes with limited clutter and does not perform as expected on our eye-in-hand robotics system, we generate more realistic synthetic training data to improve the performance. First, instead of constructing the simulation environment similar to the real world working space, we randomly place a table object imported from ShapeNet \cite{chang2015shapenet}, set the primitive shape placement area on the table, and vary the camera intrinsics and extrinsics. We adopt some of the domain randomization factors used in \cite{lin2020using}, including primitive shapes parameter, placement order, initial SE(3) object pose assigned, and mode of placement.  We also change the number of different primitive shape objects and density of placement to allow for a more cluttered environment. Furthermore, we randomly change the positions of the surrounding walls to simulate background diversity.
The network trained on the new data yields much better results, see Fig.~\ref{fig:approach_seg}.

\begin{figure}[t]
  \centering
  \begin{tikzpicture} [outer sep=0pt, inner sep=0pt]
  \scope[nodes={inner sep=0,outer sep=0}] 
  \node[anchor=north west] (a) at (0in,0in) 
    {\includegraphics[height=2.9cm,clip=true,trim=8in 2.75in 10in 4.3in]{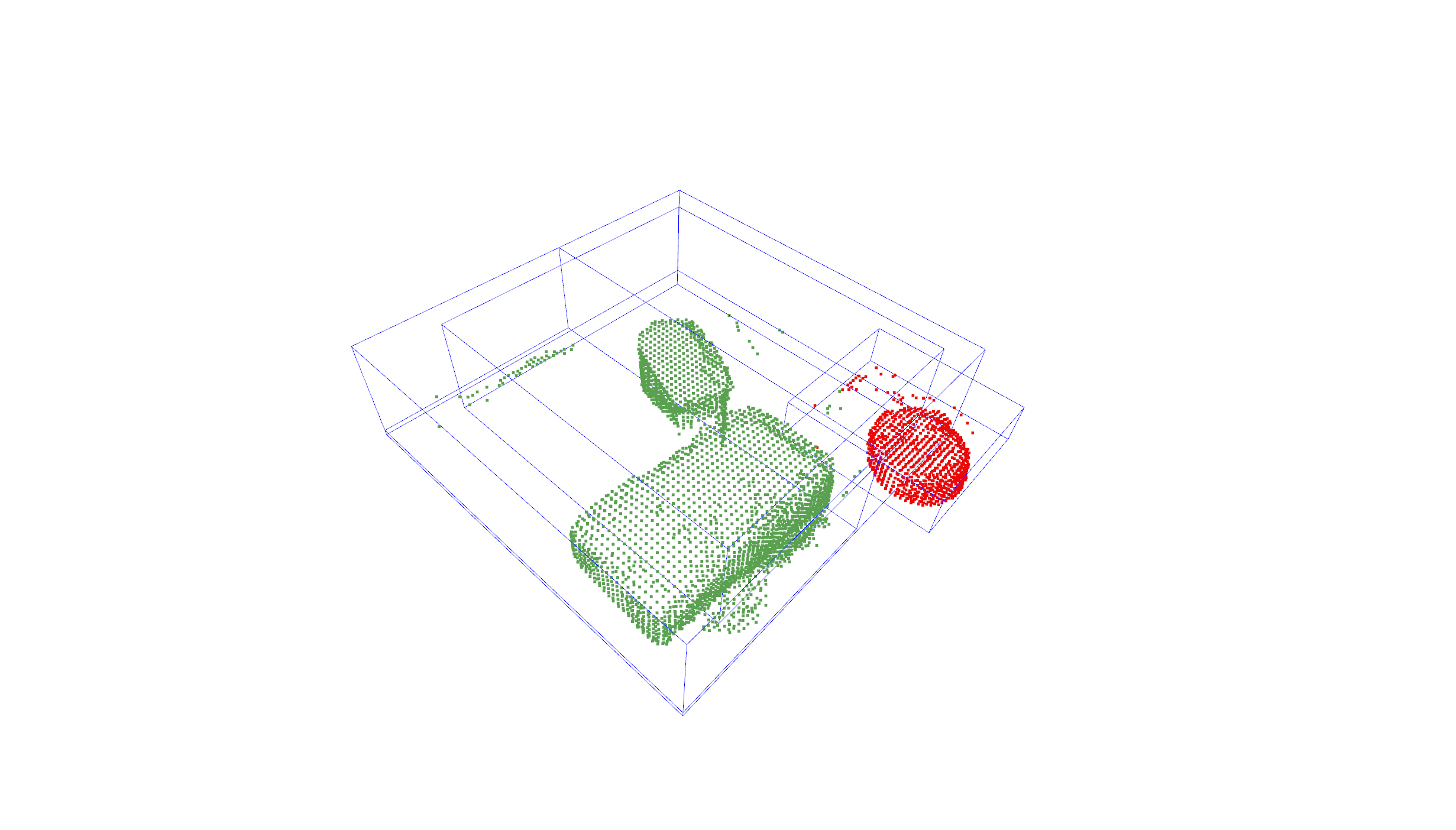}}; 
  \node[anchor=north west, xshift=0.1cm] (b) at (a.north east)
    {{\includegraphics[height=2.9cm,clip=true,trim=8in 2.75in 10in 4.3in]{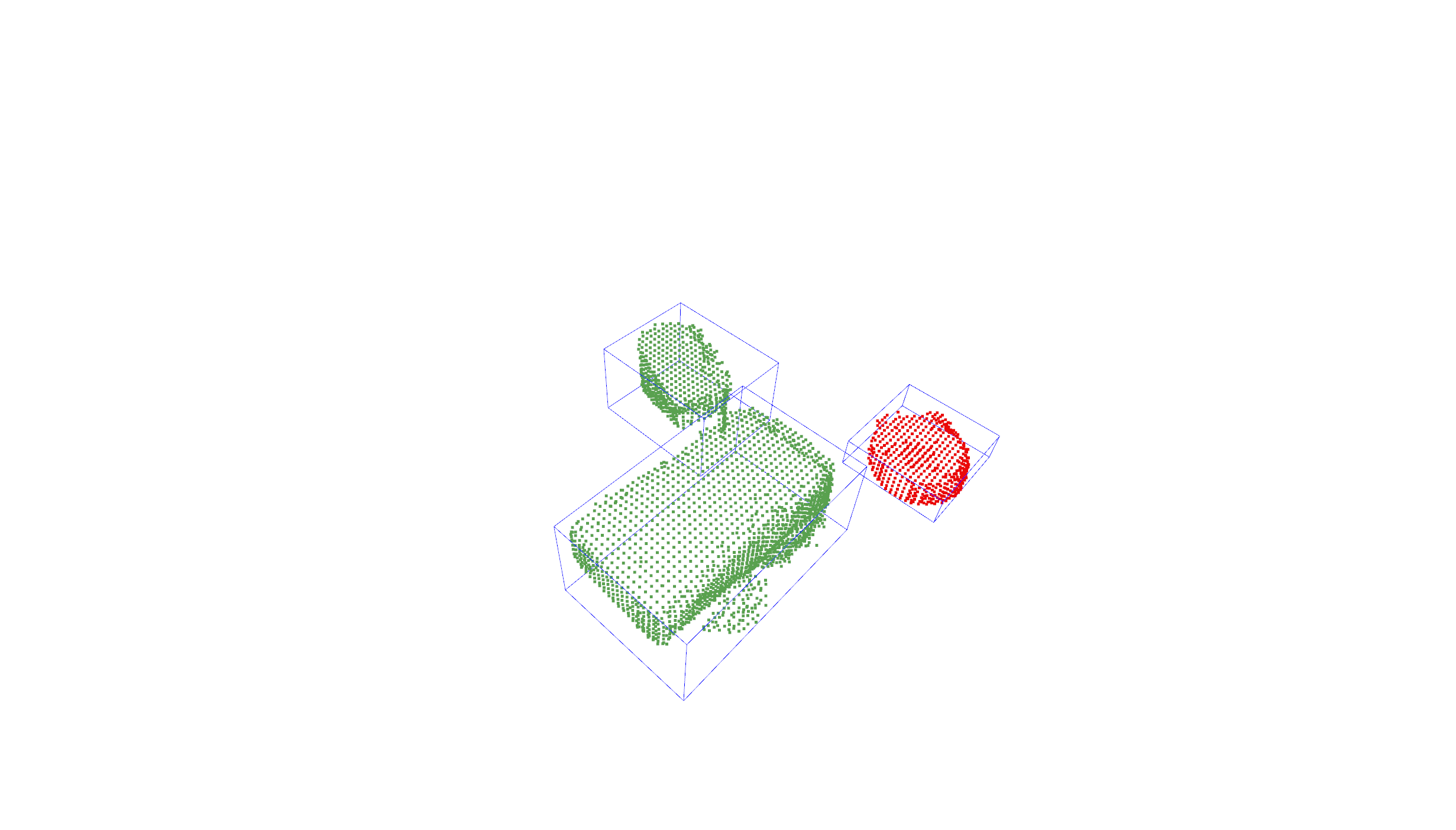}}}; 
  \endscope
  \end{tikzpicture}
  \vspace*{-1.0ex}
  \caption{Majority-voting baseline (left) yields noisier results with larger bounding boxes than our proposed multi-view voting process, which is based on a sequential aggregation strategy combined with DBSCAN and non-maximal suppression (right).}
\vspace*{-3.0ex}
  \label{fig:approach_multiview_voting}
\end{figure}

\subsubsection{Multi-view voting}

Another extension is to integrate segmentations from the newly trained network applied to multiple views.
The procedure is as follows.
The segmentations are unprojected to 3D and
voxelized, whereupon a voting process determines the voxel labels.
Based on the assumption that adjacent views would have similar predictions, point clouds corresponding to the mask instances in each view with a significant overlap to each other are combined in sequential order. After each aggregation operation, DBSCAN~\cite{ester1996density} provides denoising, along with non-maximal suppression to remove the redundant predictions according to size. 
As is shown in Fig.~\ref{fig:approach_multiview_voting}, the proposed multi-view primitive shape procedure (MV-PS) achieves better results than the majority-voting baseline, which directly calculates the majority label in a voxel across all the views.
A final RANSAC-based process fits each segmented region to a parameterized primitive shape (e.g., cylinder or cuboid) to recover a solid model representation.

\subsection{Multi-view object pose fusion}
\label{approach:pose fusion}

To retrieve the 6-DoF pose of known objects, we extend the DOPE method~\cite{tremblay2018corl:dope} to the multi-view scenario, to yield MV-DOPE.
We run DOPE on image frames captured by the robot, using a voting mechanism to merge the predictions. 
More specifically, for each object class a set $\{\bfT_i\}_{i=1}^m$ of 6-DoF poses are obtained in a common world coordinate system.
For each object pose $\bfT_i = \bmatrix{\bfR_i \,|\, \bft_i} \in SE(3)$, a confidence score $w_i^j \in \R$ is associated with the $j^\text{th}$ keypoint, 
from which the average score $w_i^{\textit{avg}} = \frac{1}{n} \sum_{j=1}^n w_i^j$ is computed, where $n$ is the number of keypoints.
Based on the assumption that a good instance candidate should have stable keypoint locations, we apply perspective-$n$-point (P$n$P) to different subsets of the keypoints to get multiple pose predictions for each detection. The consistency of the projected keypoints from these poses are then used to calculate $w_i^{\textit{pnp}}$.

\begin{figure}[t]
  \vspace*{0.075in}
  \centering
  \begin{tikzpicture} [outer sep=0pt, inner sep=0pt]
  \scope[nodes={inner sep=0,outer sep=0}] 
  \node[anchor=north west] (a) at (0in,0in) 
      {\includegraphics[width=4.2cm,clip=true,trim=2in 1.5in 2in 0.7in]{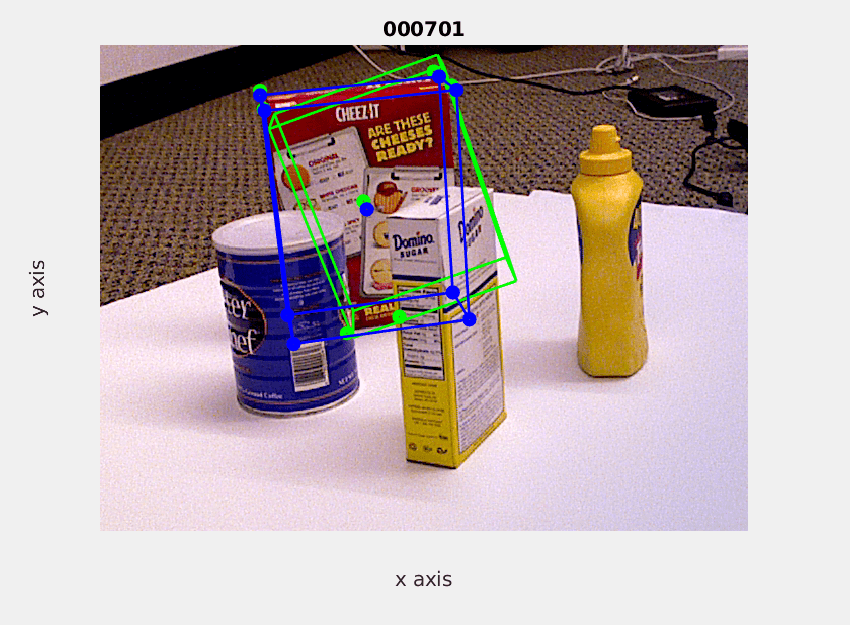}}; 
  \node[anchor=north west, xshift=0.1cm] (b) at (a.north east)
    {{\includegraphics[width=4.2cm,clip=true,trim=2in 1.5in 2in 0.7in]{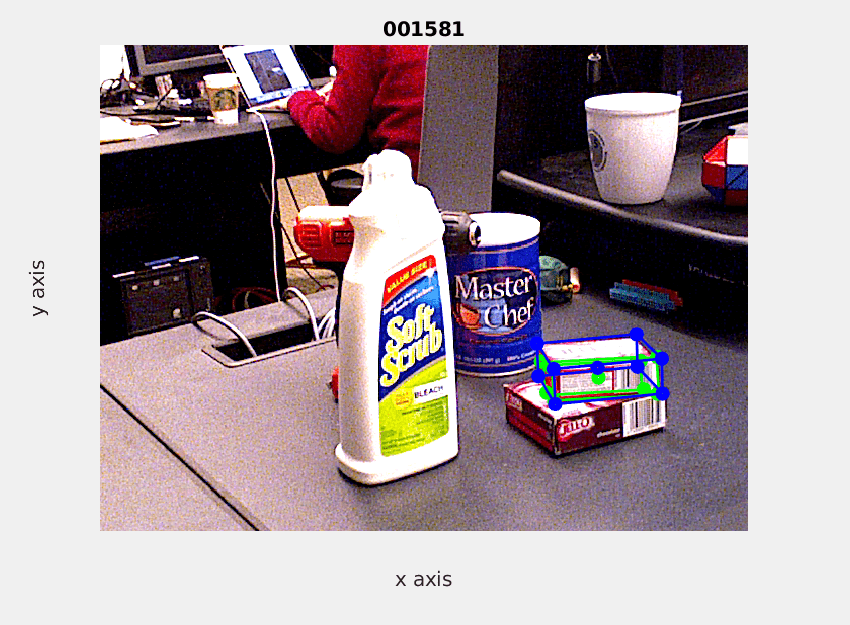}}}; 
  \node[anchor=north west, yshift=-0.1cm] (c) at (a.south west)
  {\includegraphics[width=4.2cm,clip=true,trim=2in 1.5in 2in 0.7in]{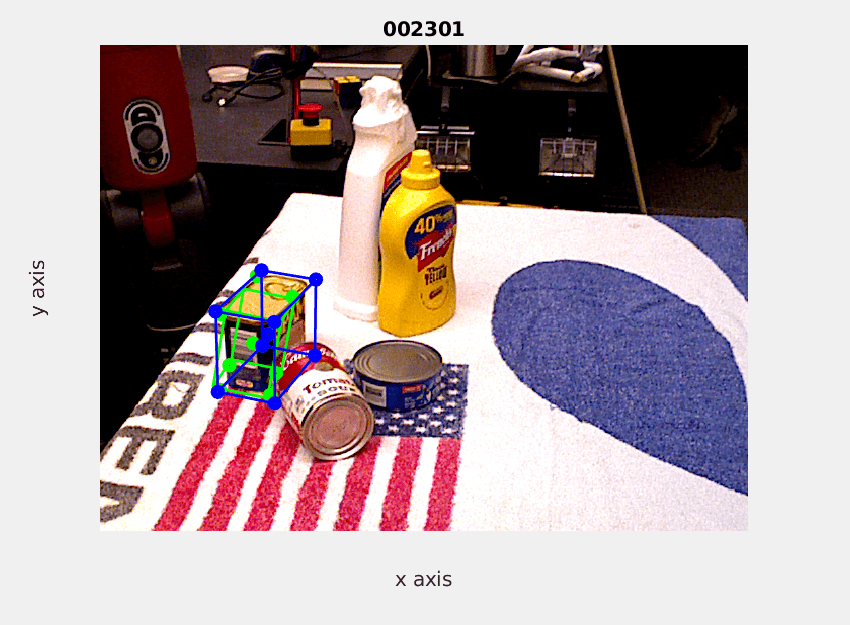}}; 
  \node[anchor=north west, xshift=0.1cm] (d) at (c.north east)
  {\includegraphics[width=4.2cm,clip=true,trim=2in 1.5in 2in 0.7in]{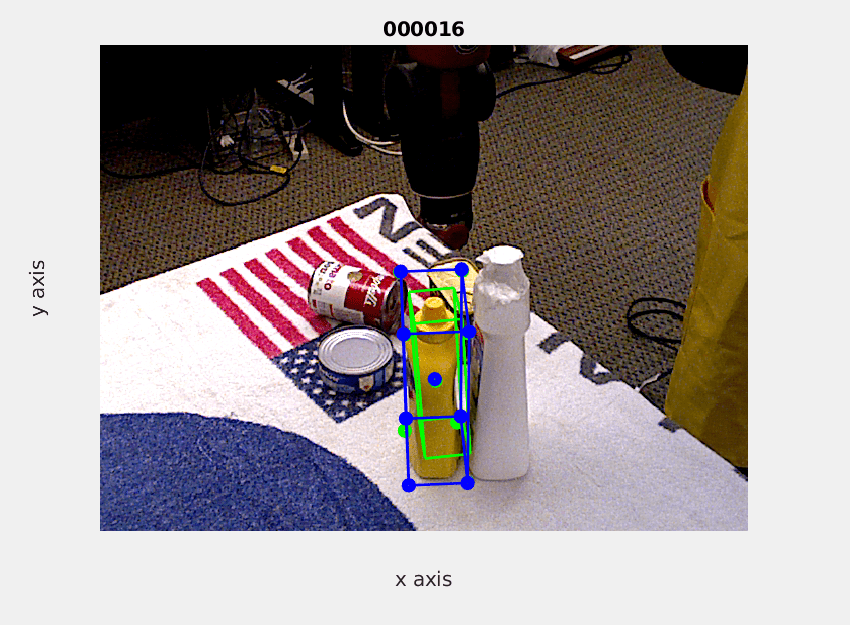}}; 
 
  \endscope

  \foreach \n in {a,b,c,d} 
   {
    \node[anchor=north west,fill=white,draw=white,inner sep=1pt] at (\n.north west) {(\n)};
   }

  \end{tikzpicture}
  \vspace*{-1.0ex}
  \caption{
  Our multi-view pose estimation (blue) is better able to handle challenging conditions than the original single-view pose estimation (green).  The images show (a) occlusion, (b) extreme lighting conditions, (c) reflective metallic surface, and (d) textureless object.  Only one object detection per image is shown to avoid unnecessary clutter.\label{fig:approach_pose_fusion}}
  \vspace*{-0.1in}
\end{figure}

Object pose candidates are filtered according to their confidence score and  Euclidean distance to different predictions.
Candidate poses are then sampled around the detected rotations $\bfR_i$ using a Gaussian, while keeping the positions $\bft_i$ fixed.  This generates a set $\calT$ of candidate poses.  The best candidate is found by minimizing the sum of weighted reprojection errors of the keypoints across all candidates~\cite{collet2010efficient}:%
\begin{equation}
\bfT^*=\underset{\bfT \in \calT}{\arg \min } \sum_{i=1}^{m} \sum_{j=1}^{n} {\tilde w}_i^j\left[\operatorname{proj}\left(\bfT \bfk^{j}\right)-\operatorname{proj}\left(\bfT_{i} \bfk^{j}\right)\right]^{2},
\label{equ:reprojection}
\end{equation}
where $\operatorname{proj}$ represents the projection operation, $\bfk^j \in \R^3$ represents the $j$th keypoint on the object model, and ${\tilde w}_i^j = w_i^{pnp} w_i^{avg} w_i^j$.

Finally, the weights are updated by comparing the detected rotations, after clustering via X-means \cite{pelleg2000x}, to those of the best candidate:  ${\tilde w}_i^j = w_i^{resample} w_i^{pnp} w_i^{avg} w_i^j$, where $w_i^{resample}$ is high when the rotation of the mean of the cluster is similar to $\bfR^*$.
These candidates are then augmented with candidate poses that are sampled around the best position $\bft^*$ and rotation $\bfR^*$ using a Gaussian with large variance to yield a new set $\calT$.
Eq.~(\ref{equ:reprojection}) is applied again with these new values to update $\bfT^*$, followed by Levenberg-Marquardt~\cite{more1978levenberg} to refine the best pose. 
Fig.~\ref{fig:approach_pose_fusion} shows the results of the multi-view DOPE and the original single-view DOPE methods.

\section{Experimental Results \label{ExpResults}}

In this section, we present our HOPE-video dataset, evaluate the three components of our system individually, and validate their integration.

\subsection{Hope-video dataset}

Considering the lack of datasets of objects (with corresponding pose estimators) on a tabletop, we introduce a dataset called ``HOPE-video'' to evaluate multi-view primitive shapes, multi-view DOPE, and the integrated system.
We collected the videos by placing a subset of the 28 HOPE objects~\cite{tremblay2020indirect} on a table in front of the robot.
The videos were captured using a 
camera mounted on the wrist of Baxter robot, providing $640 \times 480$ RGB images at 30~fps. 
We applied COLMAP~\cite{schonberger2016structure} to refine the camera poses (keyframes at 6~fps) provided by forward kinematics and RGB calibration from the camera to Baxter's wrist camera. 
We generated a 3D dense point cloud via CasMVSNet~\cite{gu2020cascade}. 
Ground truth poses for the HOPE object models in the world coordinate system were annotated manually. 
The dataset consists of 2038 images (10 videos) with 5--20 objects on a tabletop scene.
An additional 5 videos consist of a mixture of HOPE objects and unknown objects (\emph{i.e.}, objects without CAD models).

\subsection{Multi-view stereo for 3D dense reconstruction}
\label{exp:multi-view_stereo}
To test our hypothesis that deep-learning-based RGB reconstruction can rival RGBD methods in realistic  scenarios, we evaluate 3D reconstruction using the CoRBS~\cite{wasenmuller2016corbs} RGBD dataset. 
The dense RGB-based CasMVSNet~\cite{gu2020cascade}
approach that we are using 
achieves almost identical overall performance to
a leading RGBD-based algorithm.
For more details, see the Appendix.

\subsection{Multi-view primitive shape segmentation and fitting}
\label{exp:multi-view_primitive}
Following \cite{lin2020using}, we trained Mask R-CNN \cite{he2017mask} on 75k simulated depth images corrupted by a region-specific oil-painting filter.
ResNet-50-FPN was used as backbone, with 4 images per mini-batch. 
Training involved 100k iterations with initial learning rate set to 0.02 and divided by 10 at iterations 30k, 50k, and 80k.  Synthetic data generation required
$\sim$48 hours, and training required $\sim$24 hours.
All experiments were run on an NVIDIA GTX 1080Ti. 
We compared this trained network (MV-PS) with PS-CNN~\cite{lin2020using} using the test sequences of 
YCB-video~\cite{xiang2018rss:posecnn} and our own HOPE-video dataset.

We employ the ADD-S metric~\cite{xiang2018rss:posecnn} to compare the estimated pose with ground truth for each model. 
To overcome the incompleteness of the 3D dense point cloud introduced by the limited camera views, we first discard the vertices in the ground truth mesh outside a small epsilon of the point cloud. 
Then we assign segmentations to the nearest ground truth, treating multiple assignments as false positives while unmatched ground truths are considered as false negatives.

\begin{figure}[t]
  \vspace*{0.075in}
  \centering
  \begin{tikzpicture} [outer sep=0pt, inner sep=0pt]
  \scope[nodes={inner sep=0,outer sep=0}] 
  \node[anchor=north west] (a) at (0in,0in) 
    {\includegraphics[height=3.3cm,clip=true,trim=2in 0.5in 2in 2in]{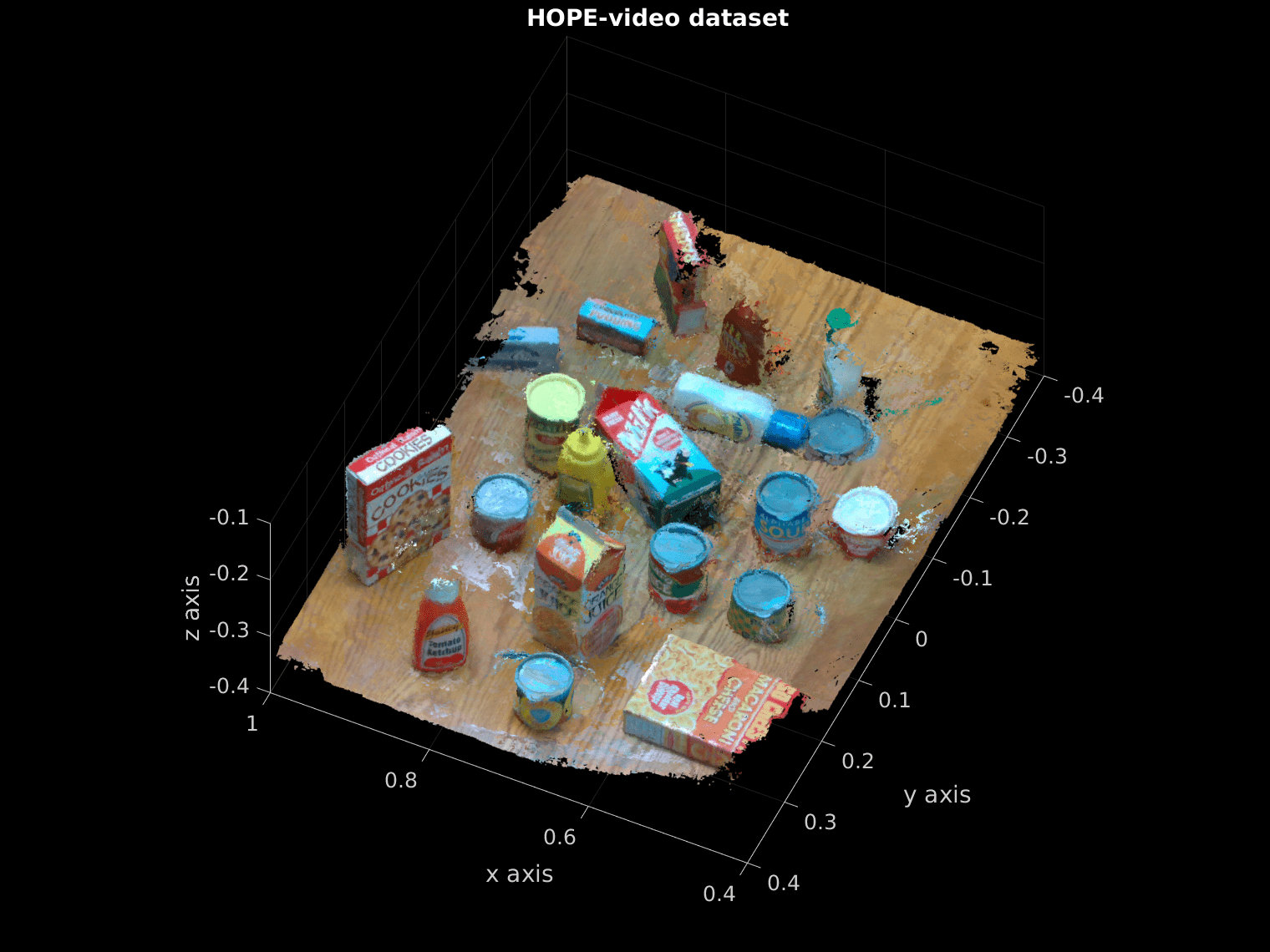}}; 
  \node[anchor=north west, xshift=0.1cm] (b) at (a.north east)
    {{\includegraphics[height=3.3cm,clip=true,trim=2in 0.5in 2in 2in]{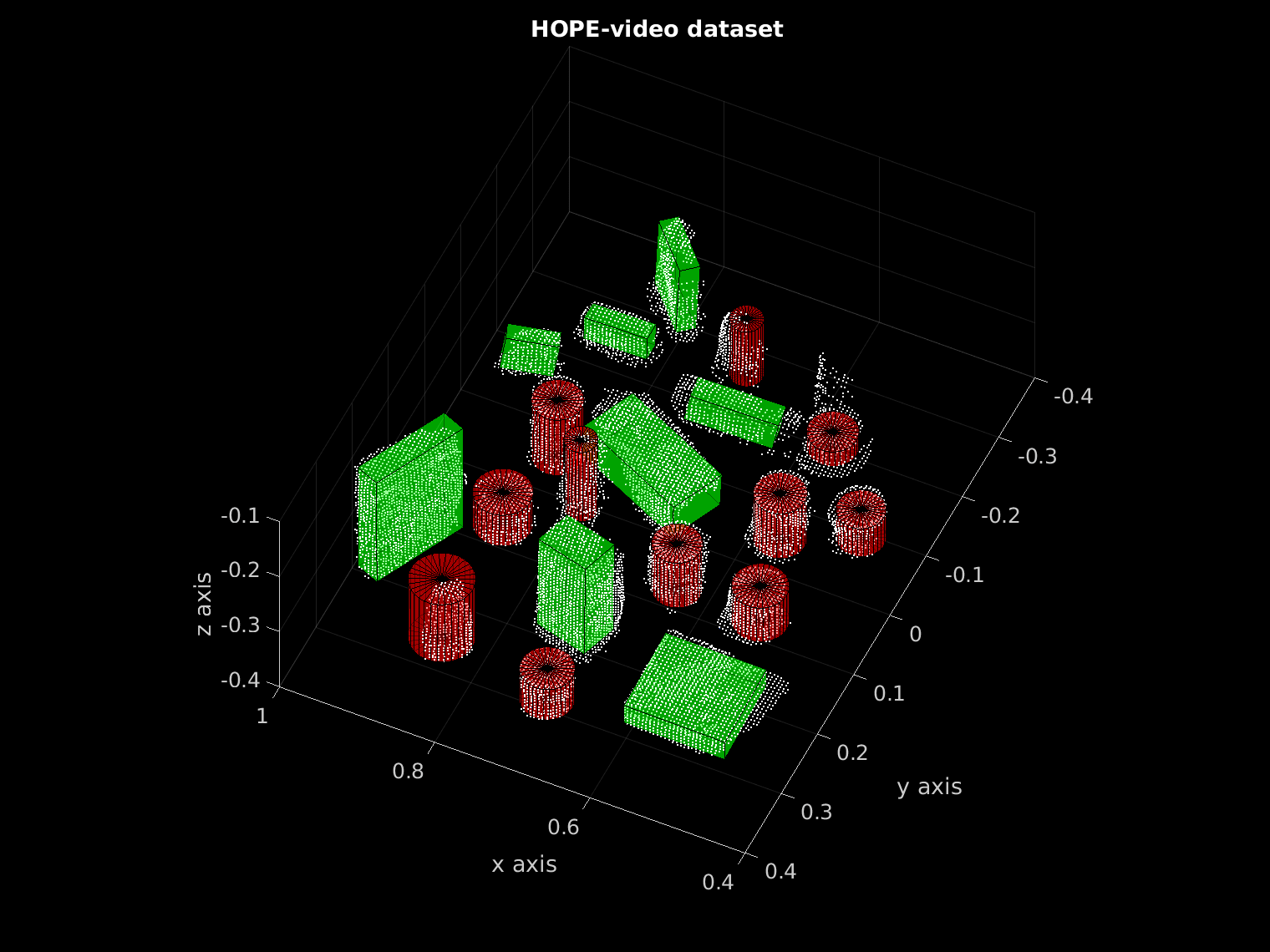}}}; 
  \endscope

  \end{tikzpicture}
  \vspace*{-1.0ex}
  \caption{{\sc Left:} 3D point cloud from CasMVSNet~\cite{gu2020cascade} on a sequence from our HOPE-Video dataset. {\sc Right:}  The final shape fitting result of the proposed method, where 19/20 of the objects are successfully recovered.  Green indicates cuboids, while red indicates cylinders.
  }
  \label{fig:exp_primitiveshapes}
\end{figure}

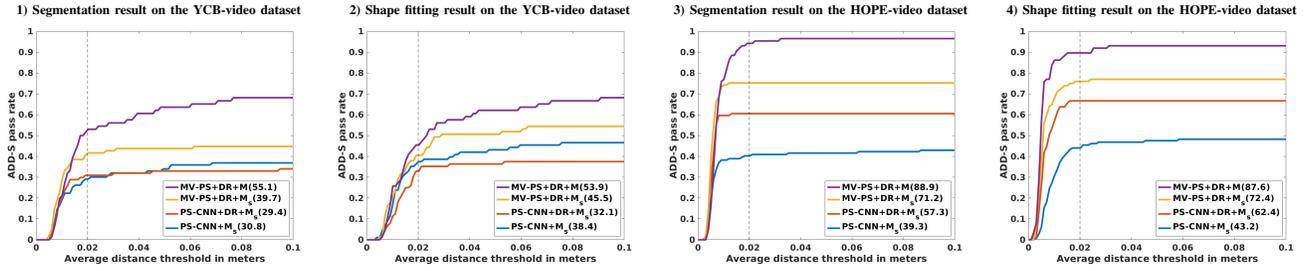
\begin{figure*}[t]
  \centering
  \begin{tikzpicture} [outer sep=0pt, inner sep=0pt]
  \scope[nodes={inner sep=0,outer sep=0}] 
  \node[anchor=north west] (a) at (0in,0in) 
    {\includegraphics[width=4.4cm,clip=true,trim=0in 0in 0in 0in]{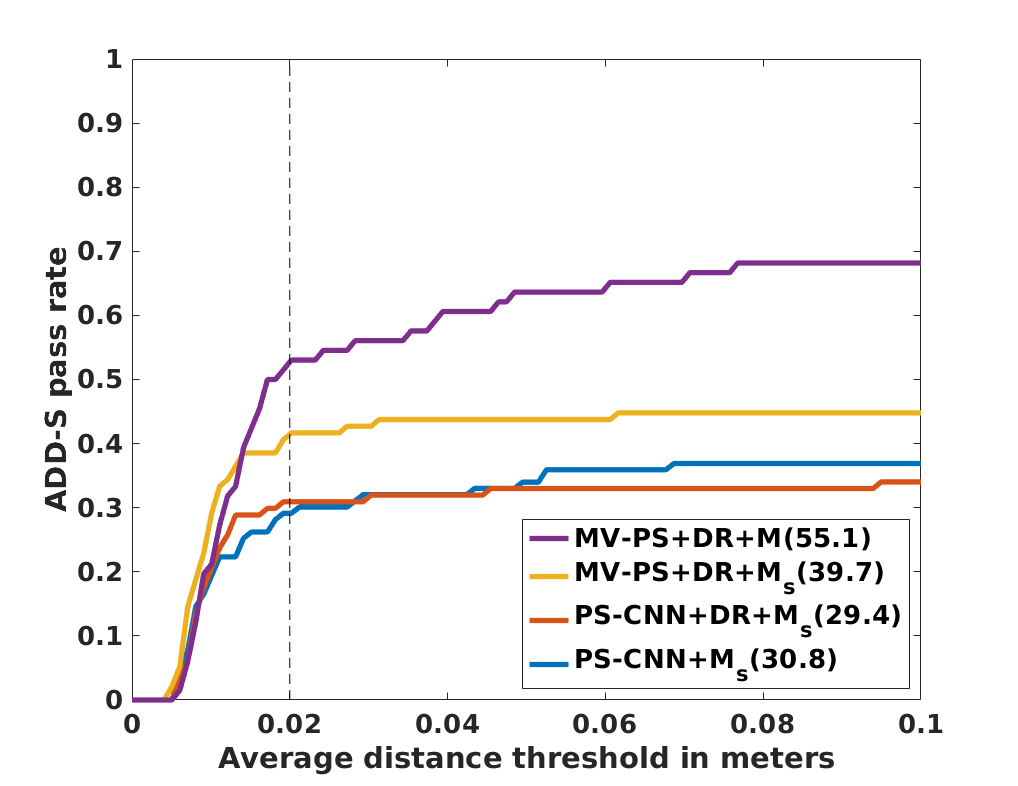}};
    \node[anchor=north, yshift=0.1cm] at (a.north) 
	    {\tiny\bf 1) Segmentation result on the YCB-video dataset};
	    
  \node[anchor=north west, xshift=0cm] (b) at (a.north east)
    {\includegraphics[width=4.4cm,clip=true,trim=0in 0in 0in 0in]{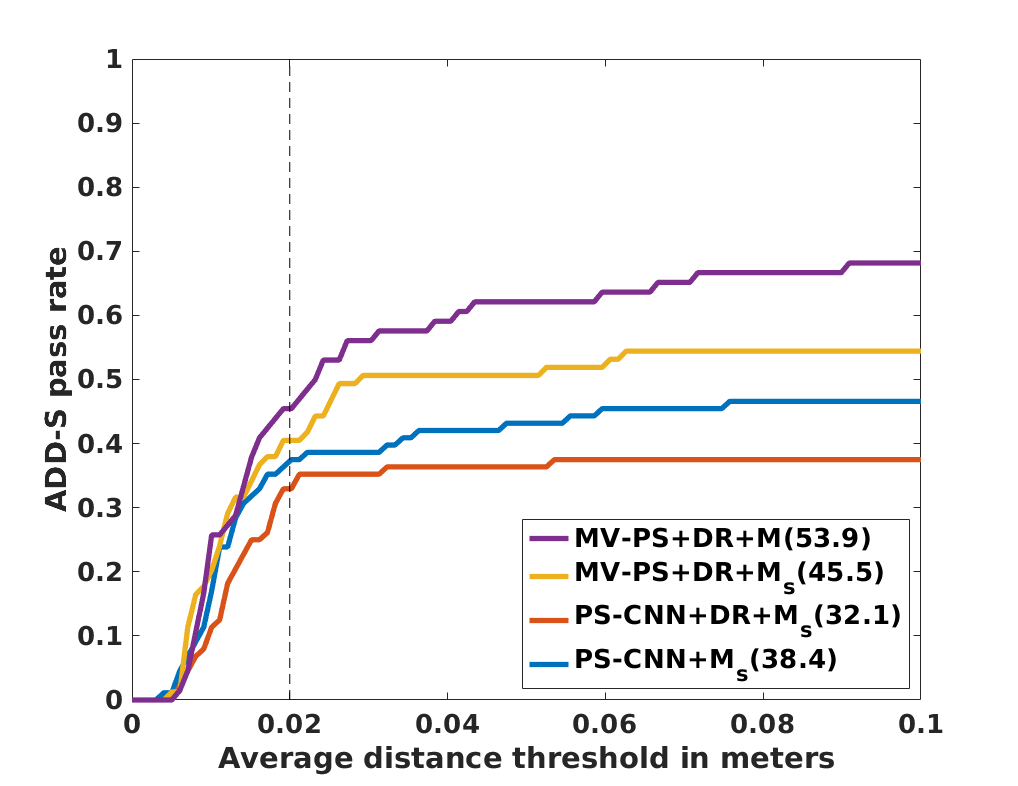}}; 
    \node[anchor=north, yshift=0.1cm] at (b.north) 
	    {\tiny\bf 2) Shape fitting result on the YCB-video dataset};
	    
  \node[anchor=north west] (c) at (b.north east)
  {\includegraphics[width=4.4cm,clip=true,trim=0in 0in 0in 0in]{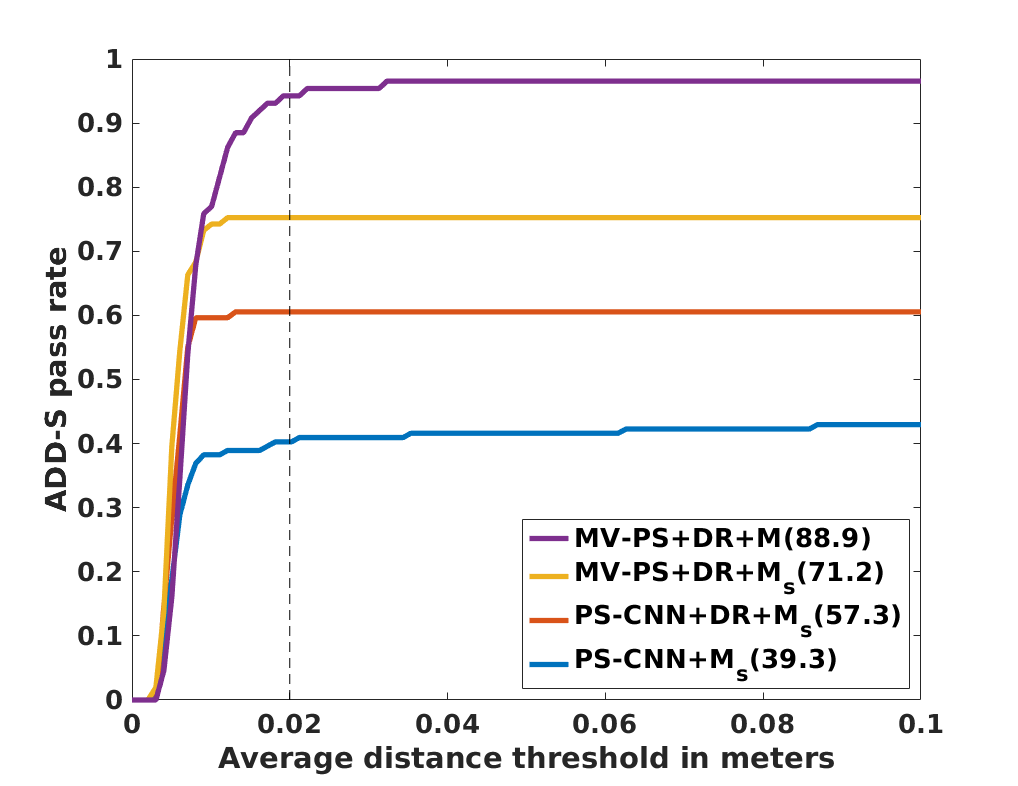}}; 
  \node[anchor=north, yshift=0.1cm ] at (c.north) 
	    {\tiny\bf 3) Segmentation result on the HOPE-video dataset};
	    
  \node[anchor=north west, xshift=0cm] (d) at (c.north east)
  {\includegraphics[width=4.4cm,clip=true,trim=0in 0in 0in 0in]{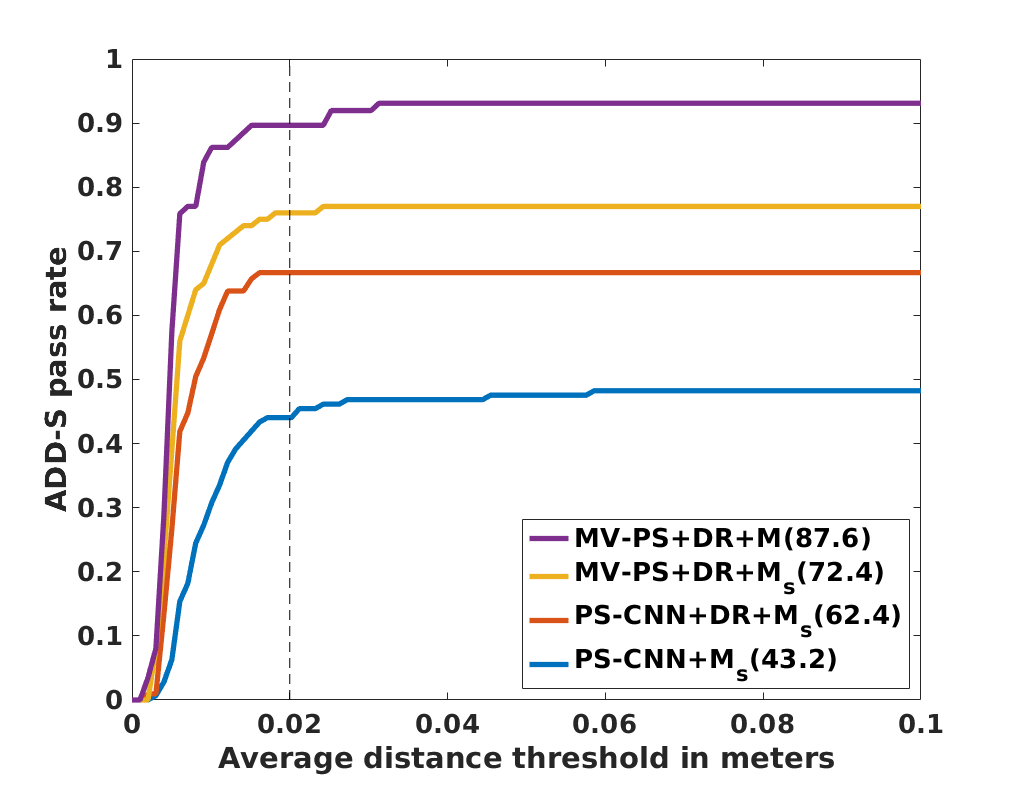}}; 
   \node[anchor=north, yshift=0.1cm ] at (d.north) 
	    {\tiny\bf 4) Shape fitting result on the HOPE-video dataset};
  \endscope

  \end{tikzpicture}
  \vspace*{-1.0ex}
  \caption{ Accuracy-threshold curves of our method (MV-PS) {\em vs.}~the baseline segmentation network PS-CNN~\cite{lin2020using}.  
  DR denotes depth refinement step, $\rm M_{s}$ is a simple multi-view integration step based on the majority-voting baseline, and M is our proposed multi-view voting method. The number in parentheses is the area under the curve (AUC).   
  Our proposed system (purple curve) outperforms the baseline and variants.}
  \label{fig:exp_primitiveshapes_AUC}
\end{figure*}

Results from our proposed segmentation network and primitive shape fitting procedure are shown in Fig.~\ref{fig:exp_primitiveshapes}.
Quantitative results on both the YCB-video and our HOPE-video datasets are shown in Fig.~\ref{fig:exp_primitiveshapes_AUC}, for both segmentation and shape fitting.
For fair comparison, we add a simple multi-view integration module based on the majority-voting idea to the baseline method \cite{lin2020using}. It directly calculates the majority label within a voxel by grouping points from all the views. 
Note that shape fitting experiments do not include filtering of ground truth vertices.
As shown in Fig. \ref{fig:exp_primitiveshapes_AUC}, the proposed method improves upon the baseline in all cases.

\begin{figure*}[ht!]
  \centering
  \scalebox{1}{
  \begin{tikzpicture}[inner sep = 0pt, outer sep = 0pt]
    \node[anchor=south west] at (0in,0in)
      {{\includegraphics[width=1\textwidth,clip=true,trim=3in 0in 3in 0in]{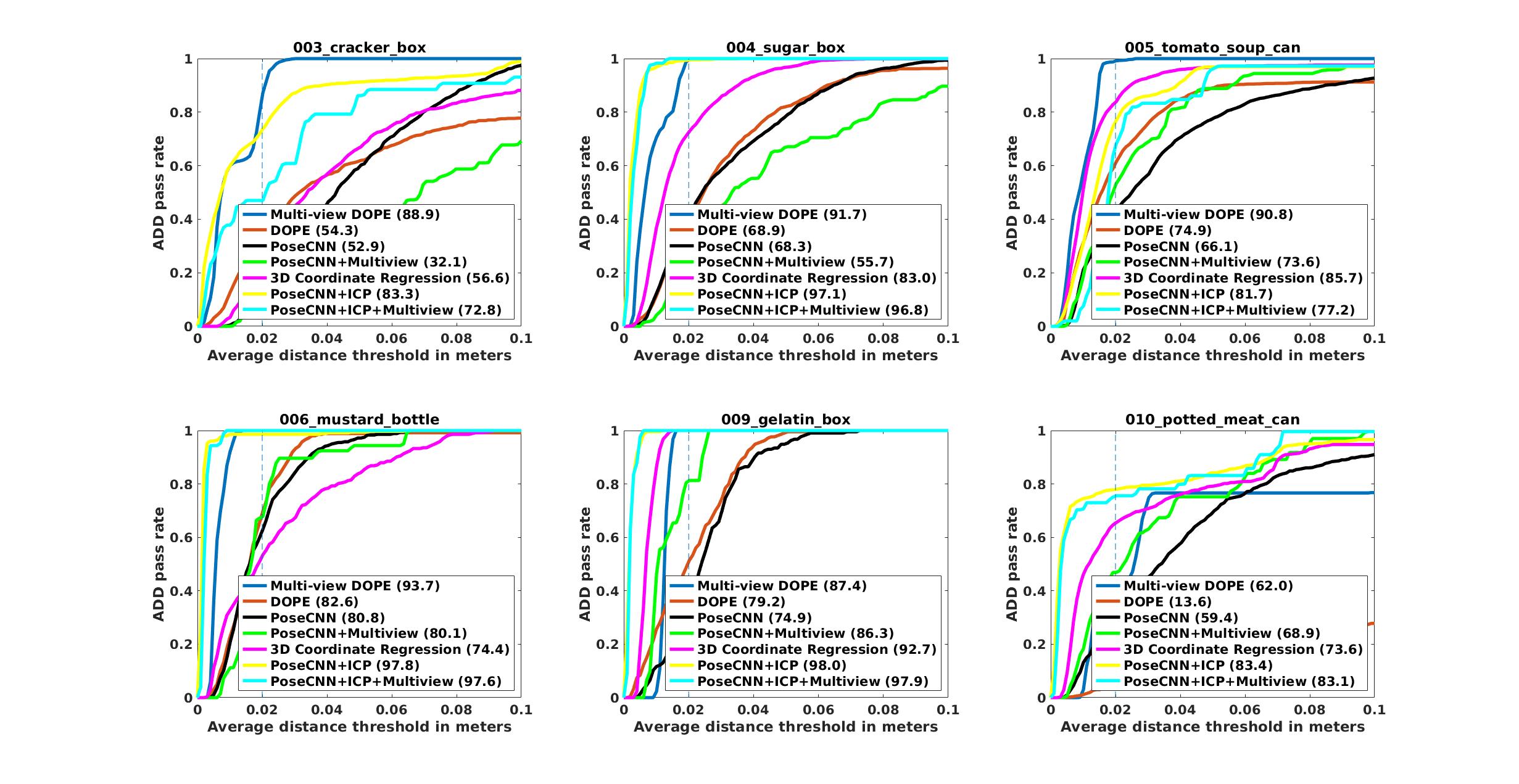}}};
  \end{tikzpicture}}
  \vspace*{-.175in}
  \caption{{Accuracy-threshold curves for our proposed multi-view fusion of DOPE compared with various single- and multi-view RGB- and RGBD-based methods. Legend numbers indicate the area under the curve (AUC). The vertical dashed line indicates the approximate grasping threshold (2 cm).}
    \label{fig:exp_AUC}}
    \vspace*{-3.0ex}
\end{figure*}

\begin{figure}[t]
    \centering
    {\includegraphics[height=6cm,clip=true,trim=0in 0in 0in 0in]{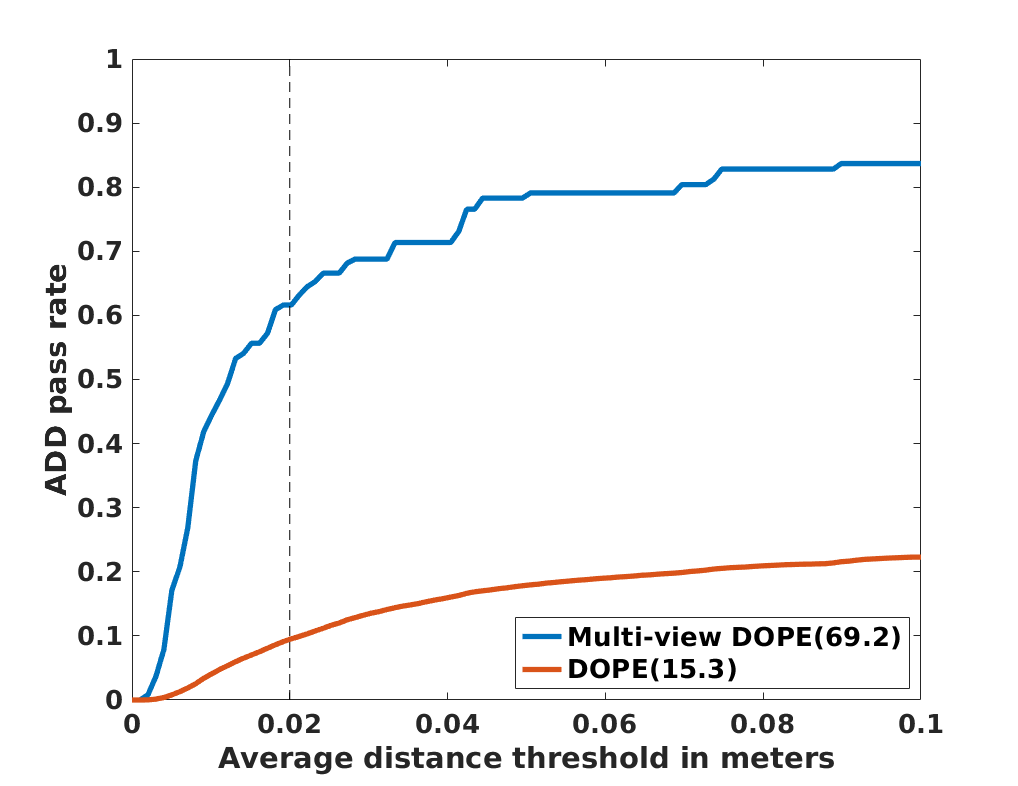}}
    \caption{ Accuracy-threshold curves for our proposed multi-view DOPE compared with the original single-view DOPE \cite{tremblay2018corl:dope} on all the HOPE objects in the HOPE-video dataset.
    \label{fig:AUC_hope_video}}
\end{figure}

\subsection{Multi-view object pose fusion}
\label{exp:multi-view_pose_fusion}
We tested our multi-view object pose estimation fusion approach also using both YCB-video and HOPE-video.
Specifically, we tested using the same 6 YCB objects and 2949 test frames as DOPE~\cite{tremblay2018corl:dope} on the YCB-Video dataset, namely,
cracker box (003), sugar box (004), tomato soup can (005), mustard bottle (006), gelatin box (009), and potted meat can (010). 
Since all models are asymmetric, we use the ADD metric~\cite{hinterstoisser2012model}, which is the average distance between the corresponding 3D points on the object model at ground truth and estimated poses. 
For a fair comparison, we use the weights publicly available online and keep the default parameters. 
For each video sequence, we retain 1 in 20 frames for computational efficiency. For HOPE-video, we test on all the 28 HOPE objects shown in the scene on all the keyframes.

In the first step, we sample 20 times around $\bfR_i$ of each candidate with a Gaussian, $\sigma=0.001$. 
In the second step, we sample 100 times around $\bft_i$ of each candidate with a Gaussian, $\sigma=0.25$ and 10 times around $\bfR_i$ with $\sigma=0.01$. 

Fig. \ref{fig:exp_AUC} compares our proposed method with 6 alternatives, including three RGB-based methods (namely, DOPE, PoseCNN, and a multi-view version of PoseCNN), and three RGBD-based methods (3D Coordinate Regression, PoseCNN+ICP, and PoseCNN+ICP+Multiview). All the reported numbers from the variants of PoseCNN~\cite{xiang2018rss:posecnn} are publicly available online.
Our method significantly improves the accuracy of DOPE, and it outperforms other RGB methods in 5 of 6 objects. 
Results on the potted meat were lower because the training data for DOPE makes it sensitive to reflection from
metallic surfaces. 
We also show the comparison between the proposed multi-view method with single-view DOPE on the HOPE-video dataset in Fig.~\ref{fig:AUC_hope_video}, showing significant improvement using multi-view.

\begin{table}[t]
\centering
\caption{ Object detection rate on three video datasets
\label{tab:eyeball}}

\resizebox{\columnwidth}{!}{ 
\begin{tabular}{ccccc}
\toprule
                       & Objects                  & MV-PS  & MV-DOPE & Integrated      \\
\midrule
YCB-video (12 videos)  & 55                    & 63.6\% & 29.1\%  & \textbf{72.7\%} \\
HOPE-video (10 videos) & 85                    & 94.1\% & 77.6\%  & \textbf{96.5\%} \\
Mixture (5 videos)     & 47                    & 78.7\% & 29.8\%  & \textbf{80.9\%} \\
\midrule
Average                  &                   & 78.8\% & 51.3\%  & \textbf{84.4\%} \\
Standard deviation                 &  & 21.8\%  & 28.9\% & \textbf{17.2\%} \\
\bottomrule
\end{tabular}

}
\end{table}

 \subsection{System integration}
 
We tested our proposed MV-PS (multi-view primitive shapes), MV-DOPE (multi-view object pose fusion), and the integrated system on the 12 test videos from the YCB-video dataset, 10 videos from the HOPE-video dataset and 5 additional videos containing a mixture of known and unknown objects---with 187 total objects across all scenes. On YCB-video, we process MV-DOPE only on the 6 models used in the previous experiment of \S\ref*{exp:multi-view_pose_fusion}, leaving 35 out of 55 objects as unknown. In HOPE-video, all 85 objects are known, with zero unknown. On the unlabeled videos, 29 objects are unknown. We count the number of detected objects with at least 50\% overlap with the point cloud. Results in Tab.~\ref{tab:eyeball} demonstrate the performance of the integrated system versus the individual modules, where the system shows higher detection rate with lower standard deviation, thus supporting our claim that the proposed multi-level system provides the robot with greater scene awareness in more cluttered/complex environments.

\section{Conclusion}
We have proposed a multi-level representation for robotic manipulation using 
multi-view RGB images. 
Using a recent 3D scene reconstruction technique, we produce a dense point cloud, useful for obstacle avoidance. 
Using this dense representation, we extend previous work for primitive shape estimation and fitting to the multi-view case.
We also propose a multi-view approach to estimate the pose of known objects with improved accuracy over single-view estimation. The integrated system provides a more complete picture of the scene. There are still some open issues to address. First, the whole system is loosely coupled and offline (around 10 min.~for an input sequence of 100 images).
Second, our approach mainly operates on tabletop scenes. 
Third, the dense 3D reconstruction quality impacts the downstream tasks.
Future work aims to explore more challenging environments (clutter, transparency, irregular poses and shapes), enhance the system design to make it more efficient, and to integrate the perception system into robotic manipulation.

\appendix

\begin{figure}[t]
  \centering
  \begin{tabular}{cc}
   
      \hspace{-0.5em}\includegraphics[width=4.2cm,clip=true,trim=2.1in 2in 1.7in 0in]{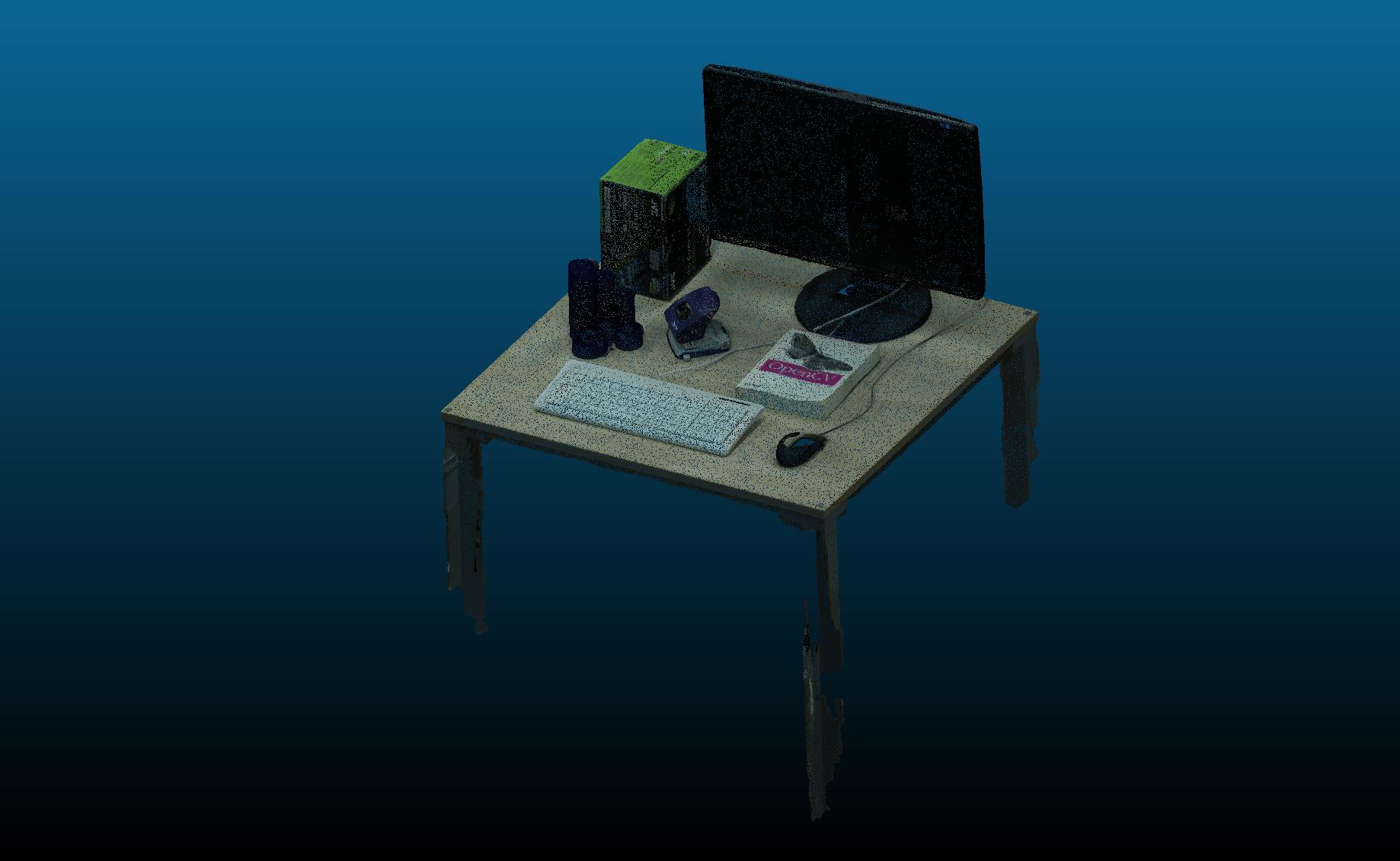} 
     &  
     \hspace{-1.0em}\includegraphics[width=4.2cm,clip=true,trim=2.5in 2in 2in 0in]{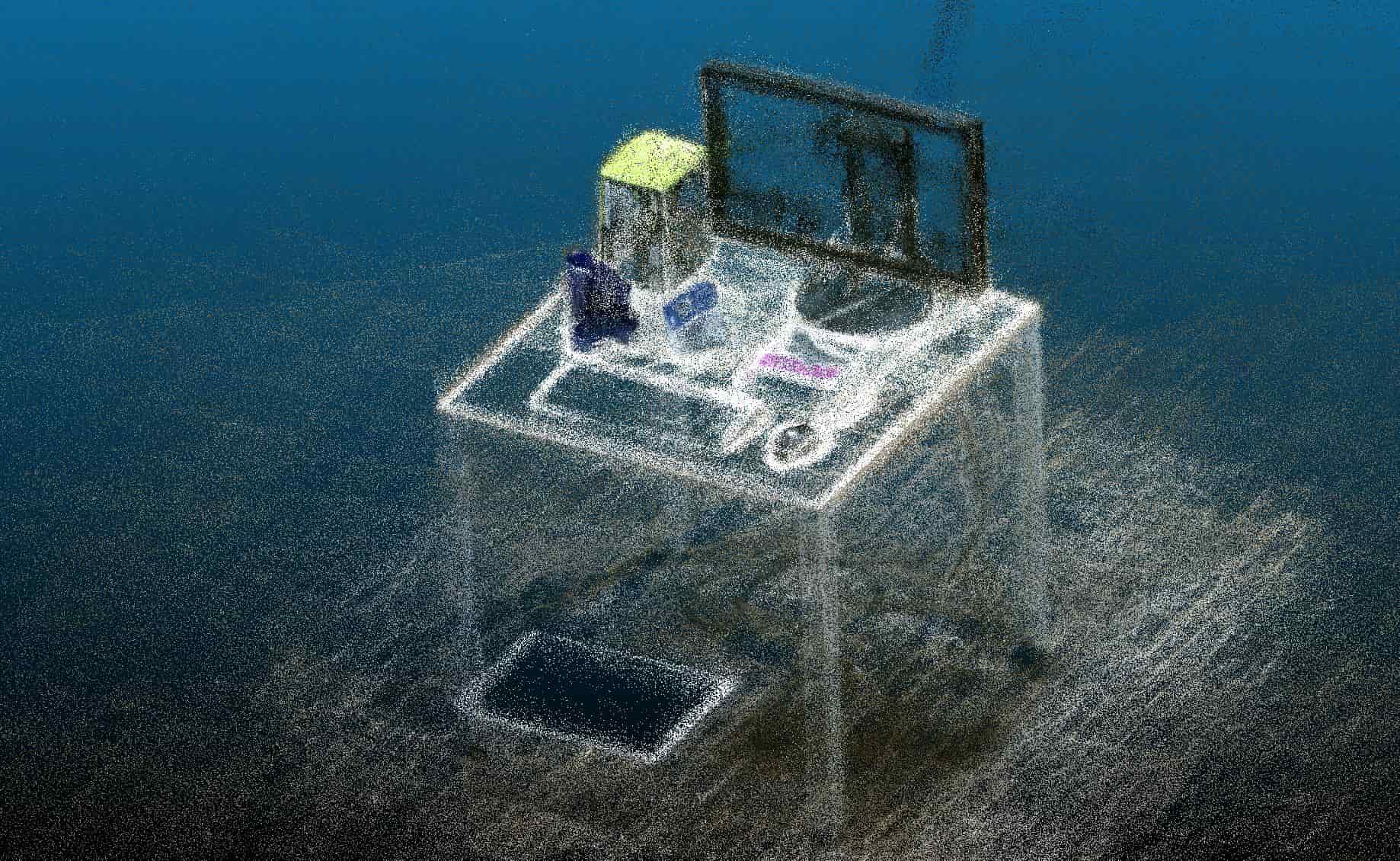} \\
     \vspace{-3ex}\raisebox{3ex}{\scriptsize{Ground truth point cloud}} &
     \raisebox{3ex}{\scriptsize{COLMAP~\cite{schonberger2016pixelwise}} (RGB)} \\
     \hspace{-0.5em}\includegraphics[width=4.2cm,clip=true,trim=2.5in 2in 2in 0in]{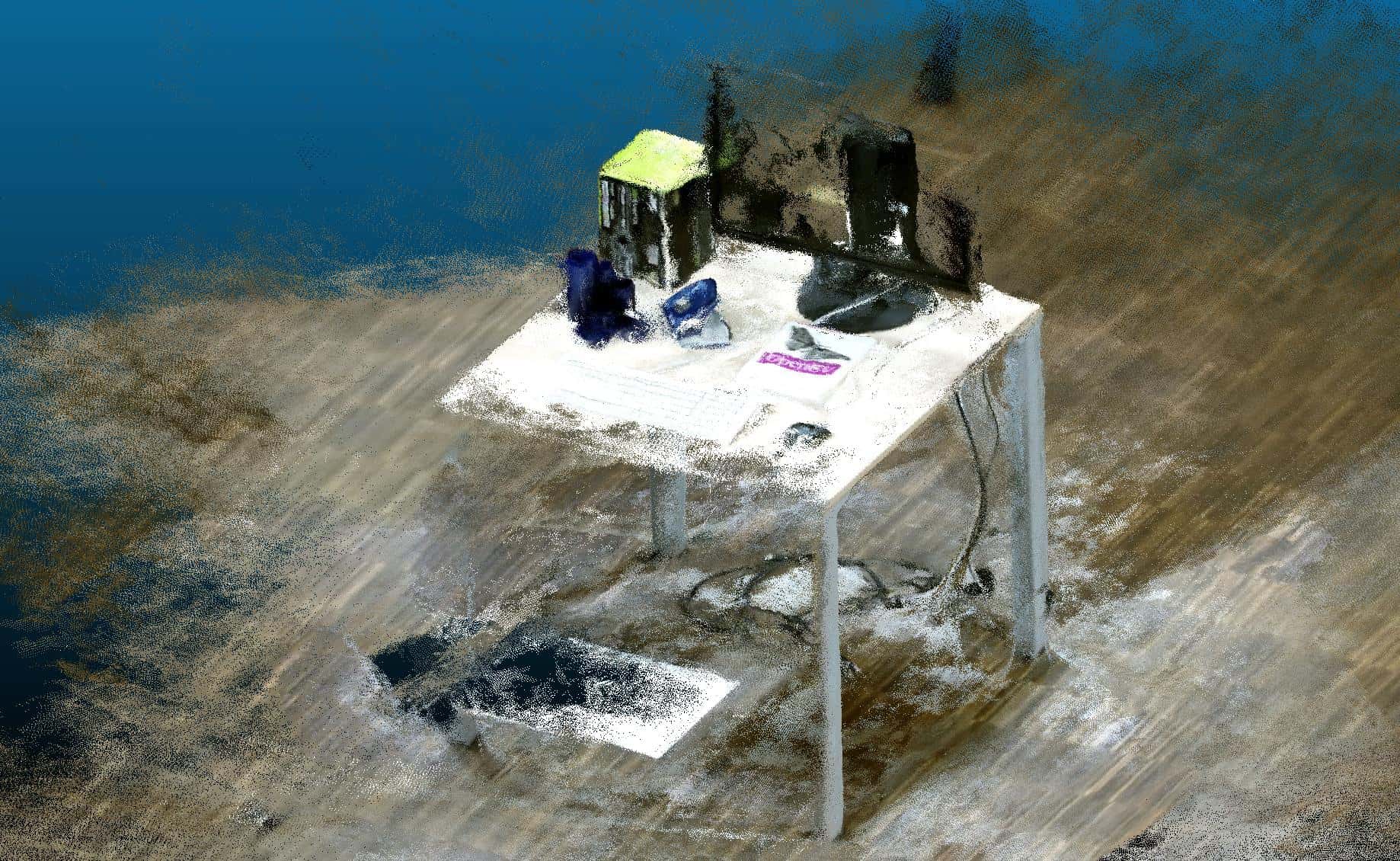} 
  & 
   \hspace{-1.0em}\includegraphics[width=4.2cm,clip=true,trim=2.5in 2in 2in 0in]{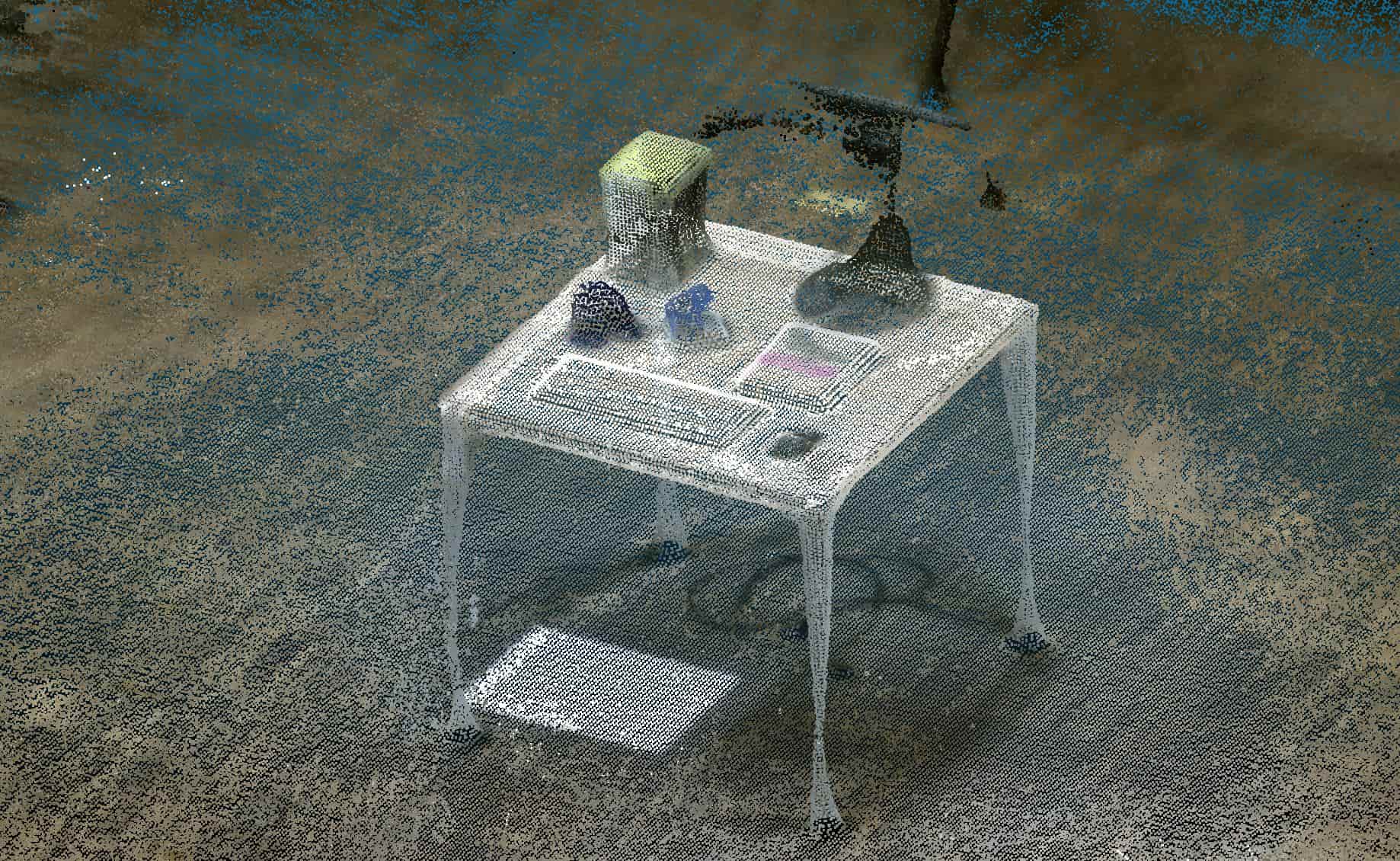} \\
   \vspace{-3ex}\raisebox{3ex}{\scriptsize{CasMVSNet~\cite{gu2020cascade}} (RGB)} &  \raisebox{3ex}{\scriptsize{Open3D~\cite{Zhou2018} (RGBD)}}
   \end{tabular}

  \vspace*{-1.0ex}
  \caption{
  Reconstruction of the CoRBS desk scene~\cite{wasenmuller2016corbs} using RGB-based methods rivals the results of RGBD-based methods.}  \label{fig:exp_3D_demo}
  \vspace*{-3.0ex}
\end{figure}

We evaluate 3D reconstruction using the CoRBS~\cite{wasenmuller2016corbs} RGBD dataset consisting of 3 scenes (desk, human, electrical cabinet)\footnote{Data for the fourth scene, racing car, is missing from the website.} and 5 trajectories for each. Images are captured by a Kinect~v2 while ground truth trajectory and scene geometry are from an external motion capture system and an external 3D scanner, respectively, each with sub-millimeter precision.  The F-score measures the accuracy and completeness of the reconstructed point clouds \cite{knapitsch2017tanks}.
For computational efficiency, image frames are temporarily subsampled, retaining 1 in 5 frames.  
For consistency with our system, COLMAP~\cite{schonberger2016structure} is applied to refine the camera trajectory, although this affects results by less than 1\% due to the high fidelity of ground truth.  The following 3D dense reconstruction methods are compared:  COLMAP, a traditional MVS method; Open3D, an RGBD integration method that utilizes depth; and CasMVSNet \cite{gu2020cascade}. For fair comparison, we use the default parameters for each method, without any postprocessing. Based on the scene dimensions, the threshold of the metric was set to 2.5, 3.5, and 2.5~cm for desk, human, and electrical cabinet, respectively. 
Tab.~\ref{tab:exp_3D} shows that CasMVSNet achieves almost identical overall performance to
Open3D, as indicated by the small gap in average scores. 
Sample results are shown in
Fig.~\ref{fig:exp_3D_demo}.

\begin{table}[t]
  \vspace*{0.075in}
  \centering
  \caption{F-score on the CoRBS dataset \cite{wasenmuller2016corbs} for 3D dense reconstruction methods \label{tab:exp_3D}}
  
\begin{tabular}{cccc}
\toprule
& COLMAP\cite{schonberger2016pixelwise}   & CasMVSNet \cite{gu2020cascade}       & Open3D \cite{Zhou2018}           \\
& RGB     & RGB              & RGBD             \\
\midrule
Desk                                                               & 68.16\% & \textbf{74.32\%} & 70.79\%          \\
Human                                                              & 83.85\% & 87.09\%          & \textbf{90.69\%} \\
Electrical cabinet                                                 & 75.30\% & 76.49\%          & \textbf{78.36\%} \\
\midrule
Average                                                            & 75.77\% & 79.30\%          & \textbf{79.95\%} \\
\bottomrule
\end{tabular}

\end{table}

\bibliographystyle{IEEEtran}
\bibliography{main.bib}

\end{document}